\documentclass[peerreviewca,12pt]{IEEEtran}
\usepackage{algorithmic}
\usepackage{algorithm}
\usepackage{amsfonts}
\usepackage{amsmath}
\usepackage{amssymb}
\usepackage{cite}
\usepackage{fancyhdr}
\usepackage{graphicx}
\usepackage{hyperref}
\usepackage{lmodern}
\usepackage[left=1in,right=1in,top=1in,bottom=1in]{geometry}
\usepackage{setspace}
\usepackage{mathtools}
\usepackage{multirow}
\usepackage{subfigure}
\usepackage{url}
\usepackage{tikz}
\usetikzlibrary{arrows,plotmarks}
\usepackage{booktabs}

\usepackage{helvet}

\makeatletter
\def\section{\@startsection{section}{1}{\z@}{-3.5ex plus -2ex minus -1.5ex}
{0.7ex plus 1ex minus 0ex}{\normalfont\large\bfseries}}
\makeatother

\newcounter{revisionCounter}
\addtocounter{revisionCounter}{1}

\newcommand\blfootnote[1]{%
  \begingroup
  \renewcommand\thefootnote{}\footnote{#1}%
  \addtocounter{footnote}{-1}%
  \endgroup
}

\title{\bf \LARGE Enhancing Selection Hyper-heuristics via Feature Transformations}
\author{

\IEEEauthorblockN{Ivan Amaya, Jos\'e C. Ortiz-Bayliss, \\ 
Alejandro Rosales-P\'erez, Andr\'es E. Guti\'errez-Rodr\'iguez, \\ 
Santiago E. Conant-Pablos and Hugo Terashima-Mar\'in, \\
School of Engineering and Sciences, Tecnologico de Monterrey, Monterrey, MEXICO
} \\

\IEEEauthorblockN{Carlos A. Coello Coello, \\
Evolutionary Computation Group, CINVESTAV-IPN, Mexico City, MEXICO
}
}


\begin{document}

\begin{figure}[t!]  
	\centering
	\includegraphics[scale = 0.60]{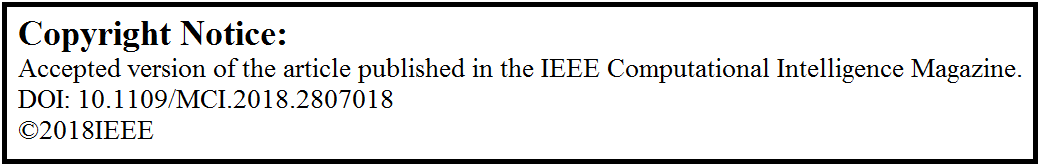}
\end{figure}

\pagestyle{empty}
\maketitle
\thispagestyle{empty}


\blfootnote{Corresponding Author: Ivan Amaya (iamaya2@itesm.mx).}

\begin{abstract}
Hyper-heuristics are a novel tool. They deal with complex optimization problems where standalone solvers exhibit varied performance. Among such a tool reside selection hyper-heuristics. By combining the strengths of each solver, this kind of hyper-heuristic offers a more robust tool. However, their effectiveness is highly dependent on the `features' used to link them with the problem that is being solved. Aiming at enhancing selection hyper-heuristics, in this paper we propose two types of transformation: explicit and implicit. The first one directly changes the distribution of critical points within the feature domain while using a Euclidean distance to measure proximity. The second one operates indirectly by preserving the distribution of critical points but changing the distance metric through a kernel function. We focus on analyzing the effect of each kind of transformation, and of their combinations. We test our ideas in the domain of constraint satisfaction problems because of their popularity and many practical applications. In this work, we compare the performance of our proposals against those of previously published data. Furthermore, we expand on previous research by increasing the number of analyzed features. 
We found that, by incorporating transformations into the model of selection hyper-heuristics, overall performance can be improved, yielding more stable results. However, combining implicit and explicit transformations was not as fruitful. Additionally, we ran some confirmatory tests on the domain of knapsack problems. Again, we observed improved stability, leading to the generation of hyper-heuristics whose profit had a standard deviation between 20\% and 30\% smaller.
\end{abstract}

\doublespace
\section{\bf Introduction}

Hyper-heuristics have emerged in recent years as a strategy for combining the strengths of different heuristics into a single method. Their aim is to provide more flexibility when solving a wider variety of optimization problems~\cite{Burke13}. Initially, authors used the term hyper-heuristic to describe heuristics for choosing heuristics.
Nowadays, this definition also includes the automatic generation of heuristics. Thus, hyper-heuristics operate at a higher level of generality by working with low-level heuristics rather than by solving a problem. Burke et al. present an in-depth survey about the topic in~\cite{Burke13}.

Selection hyper-heuristics represent a subset that relies on a mechanism for interpreting the problem state and deciding the most suitable heuristic to apply. One of the main challenges for automating this decision is how to properly characterize such a state, to allow a correct mapping to heuristics. There are different ways to get such a mapping. Examples include machine learning~\cite{ortiz2013learning} and evolutionary algorithms~\cite{Ferreira2017}. Despite this, the effectiveness of the selection depends greatly on the predictive power of the feature set~\cite{guyon2003introduction}.

Feature preprocessing is a major step in data mining and it plays a central role in the generalization performance of the models. This step encompasses selection, generation and transformation~\cite{garcia2014data}. Feature selection seeks to choose a subset of the most relevant features, for the problem at hand. Feature generation focuses on getting a new set of more discriminating features by combining primitive ones. Feature transformation aims at adapting the set of features to help learning algorithms improve their learning stage. Xue et al. offer a comprehensive review on feature selection and generation in~\cite{Xue2016, Xue2016a}. Similarly, other authors give an in-depth discussion on feature transformation in~\cite{garcia2014data, Pyle1999Book}.

In recent years, there has been an interest in studying feature preprocessing for hyper-heuristics.
Outstanding applications include~\cite{smith2012measuring, Montazeri2016, Hart2017}. In~\cite{smith2012measuring}, Smith-Miles and Lopes studied the relationships between critical features of problem instances and algorithm performance. Besides, Montazeri explored feature selection through genetic algorithms in~\cite{Montazeri2016}. Additionally, Hart et al. analyzed the effect of feature generation in hyper-heuristics~\cite{Hart2017}.
Most early studies in hyper-heuristics have focused on feature selection or feature generation while neglecting an analysis of feature transformation to empower hyper-heuristics.
The only work in this regard is~\cite{Amaya2017}, where several transforming functions were proposed. Such functions required a manual tuning of parameters. Moreover, the authors failed to assess their scalability beyond two dimensions, and its generalization for several problem domains. To the best of our knowledge, no previous study has adapted the transformation functions and explored kernel transformations~\cite{scholkopf2002learning}. Our hypothesis is that doing so may facilitate the rule-creation process (see Sect.~\ref{sec:ResultsPreliminary}). This, in turn, may allow selection hyper-heuristics to work on higher dimensional spaces and for a broader set of domains, enhancing them.

Thus, this paper makes two main contributions. First, it proposes explicit transformations, based on linear and S-shaped functions, with parameters tailored to the distribution of each feature. Second, it exchanges the distance function for one using a radial basis function kernel. Our experimental results confirm the validity of the proposed methods for improving the performance of selection hyper-heuristics in two optimization domains.

The rest of the paper is organized as follows. Section~\ref{sec:preliminaries} presents the preliminaries for properly understanding this work. We delve in the inner workings of a hyper-heuristic, and on the problems related to using features directly. Moreover, we present work previously done to explore the feasibility of using feature transformations. In this section we also present the main ideas regarding kernels and our test domain. Afterwards, we describe our proposed approach (Section~\ref{sec:ProposedApproach}), focusing on the definition of the transformations. Section~\ref{sec:Methodology} describes the experimental methodology adopted in this work, which was split into four stages. Section~\ref{sec:experiments} presents the obtained data and their discussion. Striving to facilitate the meaning of our data, we split this section into the same four stages as Section~\ref{sec:Methodology}.
Finally, the conclusions and some possible paths for future work are laid out in Section~\ref{sec:conclusions}.

\section{Preliminaries} 
\label{sec:preliminaries}

We begin this section by describing what hyper-heuristics are and how they operate. We focus on selection hyper-heuristics since it is the approach used in this work, 
although it is worth highlighting that other types of hyper-heuristics exist. 
After that, we explain two problems related to features and how transforming them may prove helpful. 
We then summarize the main idea behind a previously reported work dealing with feature transformations. 
We wrap this section up by discussing the main ideas behind a notion known as {\em Kernel} and the domain in which we carry out our experiments.

\subsection{Hyper-heuristics}

The No-Free-Lunch (NFL) theorem~\cite{Wolpert1997} implies that there is no algorithm that best solves all kinds of problems. Thus, a recent and recurring alternative is to use a strategy for combining many feasible solvers. Each of the problems that requires being solved is usually referred to as an ``instance". Even if this is a clever way to try and circumvent the restrictions posed by the NFL theorem, a recursive problem arises: the algorithm selection problem~\cite{Rice76}. Here, focus migrates to finding a proper way to carry out the selection in such a way that turns out to be beneficial. Several approaches rely on this idea to improve the generality of their solution method~\cite{OMahony2008, Malitsky2012}, but detailing them is beyond the scope of this manuscript.

\begin{figure} [ht!]
 \centering
 \includegraphics[width = 13cm]{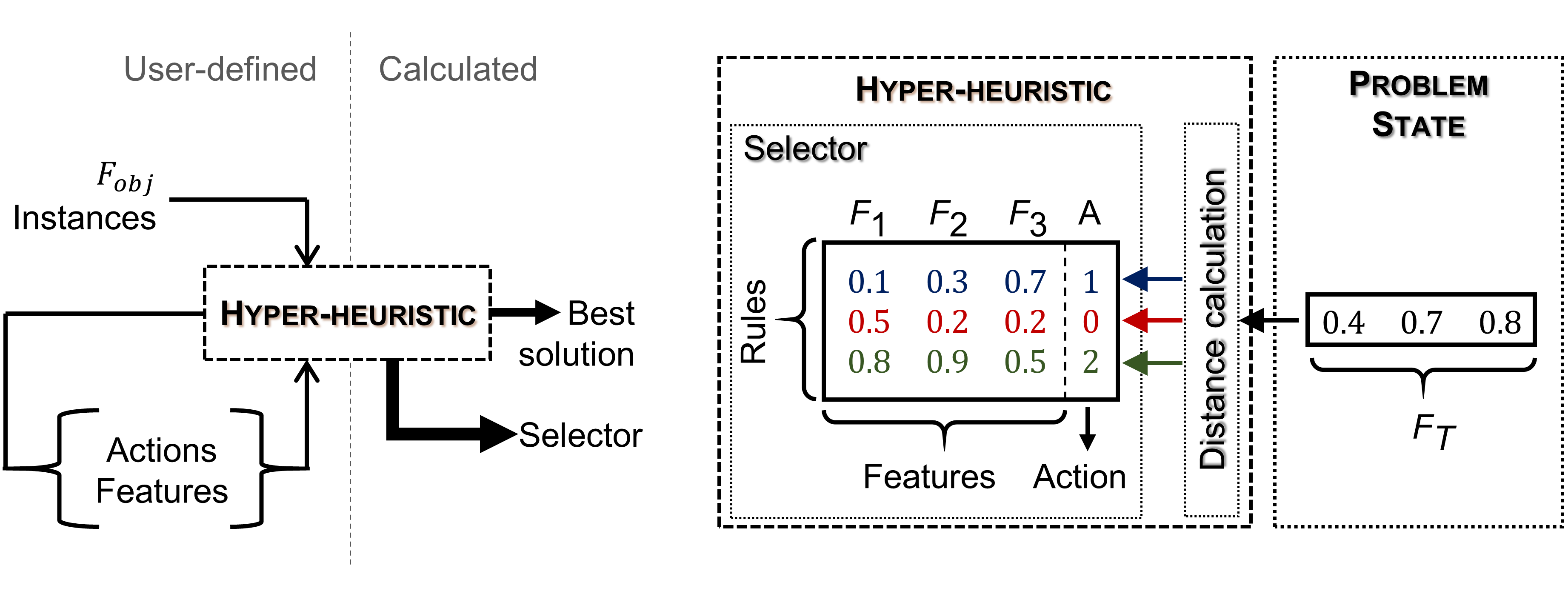}
 \caption{Overview of the hyper-heuristic model (left) and its inner workings (right). In this framework, the problem state is a vector that characterizes the instance being solved in its current stage. The hyper-heuristic contains a set of rules where the condition is represented also by a vector and their action corresponds, in this case, to specific heuristics. }
 \label{fig:HHModel}
\end{figure}

Throughout this work we use an evolutionary hyper-heuristic model proposed in~\cite{Ortiz2016}. This model falls into the category of selection hyper-heuristics, and it is depicted in Fig.~\ref{fig:HHModel}. The idea is to solve a given instance using a combination of algorithms, ruled by a ``selector". The pillar for this idea is that a partially solved problem may not be as efficiently solved by the same algorithm and, thus, a change should be made. To identify the moment in which it is appropriate to switch the solver, this model represents the problem by a set of features. These can be as simple as the size of the problem, or as complex as a relation between different values of each problem.

It is worth mentioning that the hyper-heuristic model used in this work closely resembles Learning Classifier Systems (LCS). Similarities focus on the way rules are generated and how they are applied based on the problem features. In fact, there are some works for selection hyper-heuristics that deepen into the way LCS can generate or be used as hyper-heuristics~\cite{Marin2006, Ortiz13C}.

As can be seen in Fig.~\ref{fig:HHModel} (left), the user provides four elements. The first one is the set of instances that will be solved. The second one is an objective function for measuring the quality of the solution. The remaining ones are the set of features and solvers. Using these information, the model randomly selects a subset of instances for training itself. For a given selector, the hyper-heuristic selects the first instance and calculates its features, $F_T$. With this starting point, it calculates the Euclidean distance to each rule, and applies the action (i.e., the heuristic) of the closest one. This process is depicted in Fig.~\ref{fig:HHModel} (right). After an action, the problem state changes. The process is repeated until the instance is solved. At this point, the hyper-heuristic moves to the next instance and the process continues until finishing all of them.

As mentioned before, the model used in this work has the ability to train itself. Here, this means an iterative procedure, where a messy genetic algorithm evolves a population of selectors. To do so, the process described above is carried out for each selector, and the objective function defined by the user guides the evolution. In this work, we used a steady-state configuration with a population size of 20, a crossover rate of 1.0 and a mutation rate of 0.1. Furthermore, we allowed the algorithm to run for 100 cycles (i.e., generations).

\subsubsection{An illustrative example}
Aiming to better clarify how selection hyper-heuristics work, we now present a simple, but useful, example. Imagine a problem where you have a set of items and need to split them into two subsets with the lowest possible difference. Therefore, the quality of a solution can be measured through Eq.~\ref{equ:ExampleQuality}, where $item_x^y$ represents element $x$ from subset $y$, and $N_y$ represents the number of elements in subset $y$.

\begin{equation}
\label{equ:ExampleQuality}
 Q=\left|\sum_{i=1}^{N_1}{item}_i^1 - \sum_{j=1}^{N_2}{item}_j^2\right|.
\end{equation}

One way of solving this problem requires two steps. First, assign all items to a single subset (i.e., subset 1). Then, choose some items for moving to the other subset (i.e., subset 2). Here, item transfer should stop if subset 2 represents, at least, half the items. Item selection can be done through two low-level heuristics: the Max load, and the Min load. Both can perform well or poorly, depending on the problem instance. For example, if the set of items is [10 1 1 1 1 1 1 1 1 1 1 1 1], both heuristics yield a perfect split ($Q=0$). However, if the set is [10 9 8 1 1 2 2 1 1 1 1 1 1], the quality is 15 for the Max heuristic and 1 for the Min heuristic. Moreover, if the set comprises [10 3 4 2 10 10 1 1 1 1 1 1 1], the Max heuristic yields a solution with $Q=14$ while the Min yields one with $Q=6$.

A hyper-heuristic may represent a better approach, but it requires a way of mapping features to actions (i.e., the rules). Here, a single feature ($F_1$) can be defined as shown in Eq.~\ref{equ:ExampleFeature}, and one rule can be defined for each action.

\begin{equation}
\label{equ:ExampleFeature}
 F_1 = \frac{\sum_{j=1}^{N_2}{item}_j^2}{\sum_{i=1}^{N_1}{item}_i^1+\sum_{j=1}^{N_2}{item}_j^2}.
\end{equation}

\begin{figure}[ht!]
 \centering
 \includegraphics[width = 13cm]{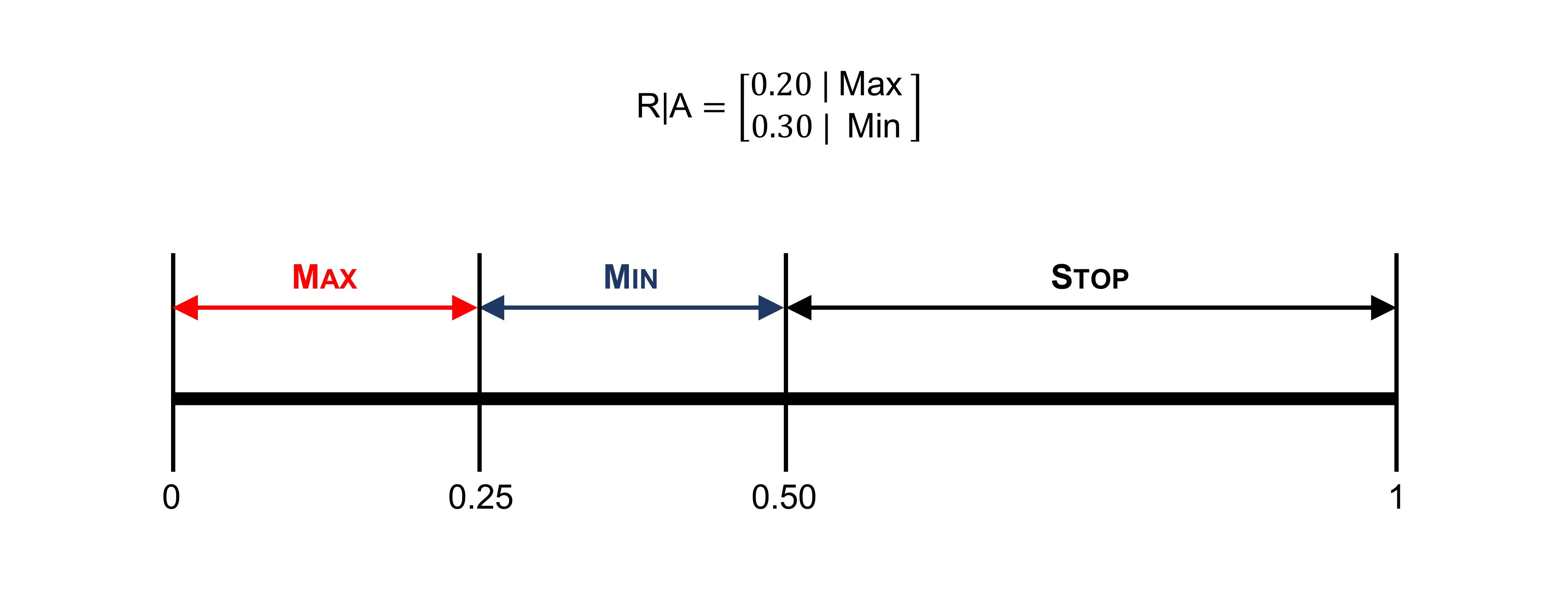}
 \caption{An example of a set of rules (top) and its corresponding zone of influence (bottom).}
 \label{fig:HH1}
\end{figure}

Figure~\ref{fig:HH1} shows a sample set of rules (top) and its corresponding zone of influence (bottom). This means that, whenever the feature value is below 0.25, an item will be moved following the Max heuristic. For higher feature values, it will be moved according to the Min heuristic.
For example, consider the last instance. The first two items will be selected using the Max heuristic, since feature values are 0 and 0.22. When selecting the third item, the feature value goes up to 0.43. Thus, from this point onward all items are selected using the Min heuristic, until the feature becomes, at least, equal to 0.5.
Through this approach, problem instances one and three can be perfectly solved (i.e., $Q=0$), while the second one can be solved with $Q=1$, thus making it a better solver.

\subsection{Problems Derived from Using Features Directly}
\begin{figure}
 \centering
 \includegraphics[width = 13.0cm]{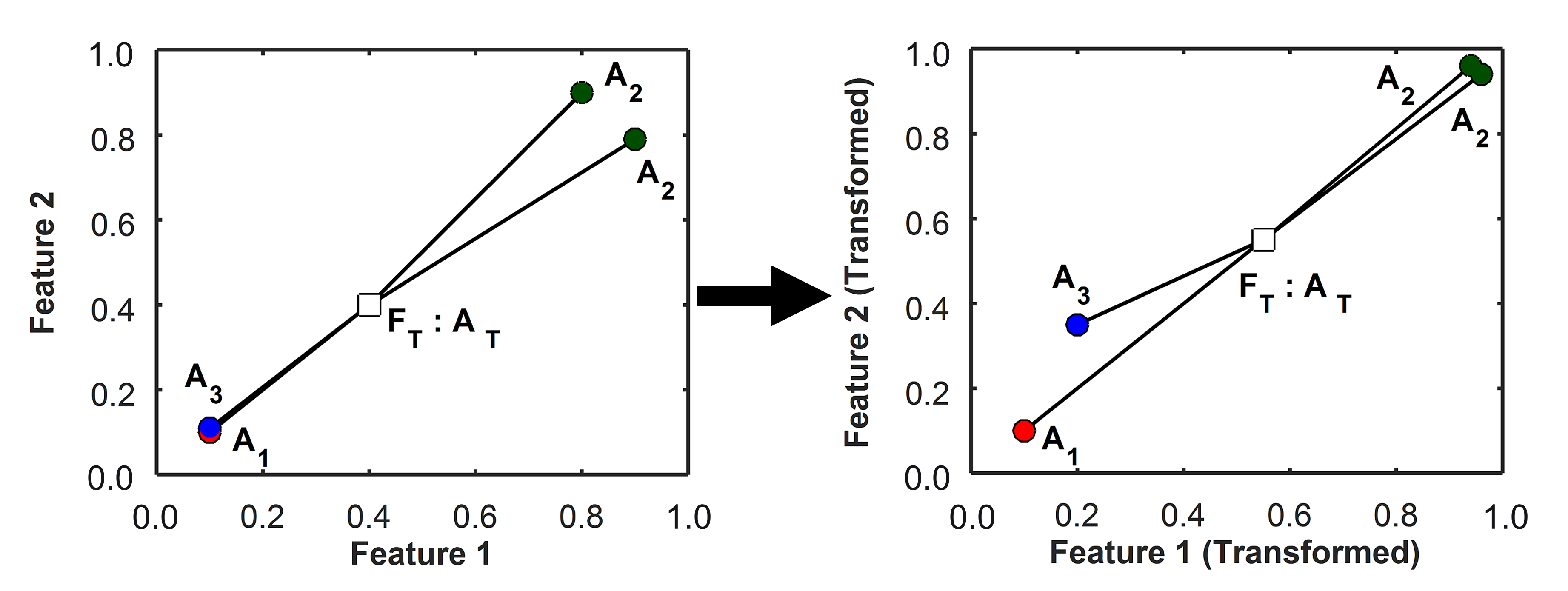}
 \caption{Distribution of rules before (left) and after (right) feature transformation.
Square: Current state of the problem. Circles: Ideal location of rules. $A_i$: Actions of each ideal rule.}
 \label{fig:PreliminariesRulesProblem}
\end{figure}

Using the aforementioned model with raw features may exhibit two drawbacks~\cite{Amaya2017}:
likeliness and stagnation. The former appears when two problem instances with similar features
are best solved by different actions (e.g., at the boundary of regions best solved by both actions).
Should one rule (from the selector) be closest to both states, one of them would not be solved in
the best possible way. Similarly, if problem states best solved by the same action are apart,
clustering them frees up space to distribute them among other actions.
Figure~\ref{fig:PreliminariesRulesProblem} shows an example of both scenarios. In the figure,
each circle represents the best location for a rule, and their corresponding actions are given
by $A_i$. The square marker represents the current state of a problem, indicated by $F_T$, and
where the action to be taken, $A_T$, must be decided by using the closest point. However, on the
left, rules one and three are so clustered that a small error when placing them (e.g., when evolving),
could lead to a wrong decision. Ot the right, transformations help by clustering alike regions
and expanding troublesome ones, allowing for a smoother change in the performance of a
selector-in-training.

The second problem, to which we refer to as stagnation, is related to the nature of optimization
procedures~\cite{Amaya2017}. During the first iterations, improvements are significant and quite
common. As the search progresses, they become less frequent and less significant. For
our hyper-heuristic model, stagnation reflects on a population of selectors with small differences in both,
features and performance. Nonetheless, by transforming features we can expand part of the feature
space, allowing for a bigger variation which may lead to improvements.

\subsection{Previous Work Related to Feature Transformations}
Feature transformation is an active research field in machine learning, where the idea is to use information from the original features to create new ones with improved predictive power. Several methods have been proposed. Some of the best known methods are normalization, standardization, and polynomial transformation~\cite{garcia2014data}. An emerging area is the use of evolutionary computation for taking into account the behavior of the recognition system~\cite{Xue2016}. In spite of 
its success, this approach is computationally costly.

\begin{figure}
 \centering
 \includegraphics[width = 13.0cm]{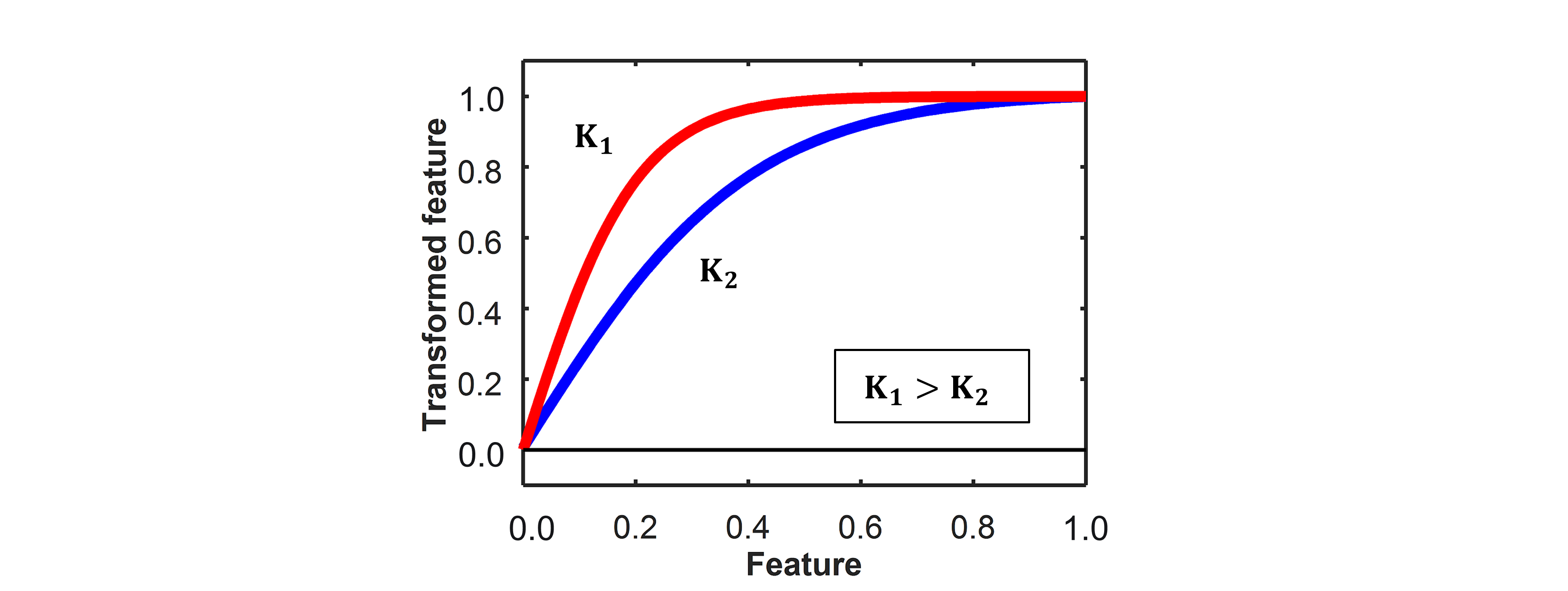}
 \caption{Overview of the exponential transformation previously used in~\cite{Amaya2017}.}
 \label{fig:FundamentalsCECTransform}
\end{figure}

We have previously studied feature transformation for improving the performance of selection hyper-heuristics. In~\cite{Amaya2017}, we applied Eq.~\ref{equ:CECTransform}, where $K=5$ is a parameter we determined empirically, and which behaves as shown in Fig.~\ref{fig:FundamentalsCECTransform}.
To improve upon that idea, in this work we test two new transformations and a way for tailoring
them to each feature (Sect.~\ref{sec:ProposedApproach}).
\begin{equation}
 \Phi(\mathbf{x}) = 1 - 2 \cdot \left( \frac{e^{-K \cdot \mathbf{x}} - e^{-K}}{1 + e^{-K \cdot \mathbf{x}}} \right).
 \label{equ:CECTransform}
\end{equation}

\subsection{Kernels}
Kernel functions implicitly embed mapping functions, and were popularized by Support Vector
Machines~\cite{Vapnik1995Book} and the so-called {\em kernel trick}. This has enabled them
to learn nonlinear functions, greatly improving their performance.

Roughly speaking, a kernel is a mathematical function that computes the inner product between
two vectors in a higher (and possibly infinite) dimensional space without explicitly performing
the mapping. A kernel function, $K$, for two points, $\mathbf{x}^{\left(1\right)}$ and
$\mathbf{x}^{\left(2\right)}$, is expressed in Eq.~\ref{eq:KernelDefinitio}, where
$\left\langle \cdot,\cdot\right\rangle$ is the inner product between two vectors.
\begin{equation}
K\left(\mathbf{x}^{\left(1\right)},\mathbf{x}^{\left(2\right)}\right) =\left\langle \Phi\left(\mathbf{x}^{\left(1\right)}\right),\Phi\left( \mathbf{x}^{\left(2\right)}\right)\right\rangle.
\label{eq:KernelDefinitio}
\end{equation}
Some common kernels are shown in Eq.~\ref{eq:LinearKernel}, Eq.~\ref{eq:PolyKernel}, and Eq.~\ref{eq:rbf_kernel}, respectively, where $\|\cdot\|$ is the Euclidean norm and $d$ and $\gamma$ are adjustable parameters.
\begin{align}
 &\mbox{Linear Kernel: } & & K_{L}\left(\mathbf{x}^{\left(1\right)},\mathbf{x}^{\left(2\right)}\right) & = & \left\langle\mathbf{x}^{\left(1\right)},\mathbf{x}^{\left(2\right)}\right\rangle & & & & & & & & & & & & & & & & & & &
 \label{eq:LinearKernel}\\
 &\text{Polynomial Kernel: } & & K_{P}\left(\mathbf{x}^{\left(1\right)},\mathbf{x}^{\left(2\right)}\right)& = & \left(\left\langle\mathbf{x}^{\left(1\right)},\mathbf{x}^{\left(2\right)}\right\rangle + 1\right)^d 
 \label{eq:PolyKernel} \\
 &\text{Radial basis function Kernel: } & & K_{RBF}\left(\mathbf{x}^{\left(1\right)},\mathbf{x}^{\left(2\right)}\right) & =  & \;\mathrm{ e}^{-\gamma\|\mathbf{x}^{\left(1\right)}-\mathbf{x}^{\left(2\right)}\|^2}.
 \label{eq:rbf_kernel}
\end{align}

\subsection{Domain Under Study: Constraint Satisfaction Problems (CSPs)}
\label{sec:CSPs}
There are many practical applications for a CSP~\cite{Berlier10,Bochkarev15}. These problems can be defined by a set of variables, $V$, where each variable $v \in V$ contains two pieces of information: its domain ($D_{v}$) and its constraints ($C$). The former represents the finite set of available values for the variable, while the latter restricts combinations of the variable with others. Solving a CSP, thus, requires assigning a feasible value to every variable (a process known as ``instantiation") in such a way that all constraints are satisfied~\cite{Amaya2017}. 

A recurring approach for solving CSPs is to use a tree representation explored in a depth-first manner. Details are omitted due to space restrictions, but interested readers are referred to~\cite{Amaya2017,Ortiz2016} for more details on this. We simply mention that every node represents the assignment of one variable. Furthermore, constraints must be checked after arriving at each node to verify the feasibility of a solution.

Eight features were considered for this work as an attempt to incorporate information from the distribution of the constraints and conflicts in the instance. The set of features is consistent with the ideas considered in~\cite{Ortiz2016}:

\begin{itemize}
 \item \textbf{Constraint density ($p_{1}$)}: The constraint density is a measure of the proportion of constraints within the instance; the closer the value of $p_{1}$ to 1, the more constraints exist within the instance.
 \item \textbf{Constraint tightness ($p_{2}$)}: The constraint tightness estimates how difficult the constraints are to be satisfied. Higher values indicate instances more likely to be unsatisfiable. The tightness of a constraint represents the proportion of conflicting tuples within such a constraint. Then, the constraint density of an instance is calculated as the average constraint tightness among all the constraints in the instance.
 \item \textbf{Clustering coefficient ($c$)}: This feature considers the instance as a graph where variables are represented as nodes and constraints as edges. The local clustering coefficient of a variable measures how close their neighbors are to being fully connected. The clustering coefficient of an instance is the average of the local clustering coefficients among all the variables.
 \item \textbf{Upper and lower constraint density quartiles ($UQp_{1}$ and $LQp_{1}$)}: These two features provide information of the distribution (based on the upper and lower quartiles) of the constraint density of the individual variables within the instance.
 \item \textbf{Upper and lower constraint tightness quartiles ($UQp_{2}$ and $LQp_{2}$)}: The same idea than in the previous feature but focused on the constraint tightness of each particular constraint.
 \item \textbf{Kappa ($\kappa$)}: This concept is suggested in the literature as a general estimation of how restricted a combinatorial problem is~\cite{Gent96B}. If $\kappa$ is small, the problems usually have many solutions to their size. When $\kappa$ is large, instead, the problems often have few solutions or have none at all.
\end{itemize}

Along with the features previously described, four commonly used heuristics are given as tools for hyper-heuristics:

\begin{itemize}
\item \textbf{DOM}: DOM instantiates first the variable that is more likely to fail. DOM estimates how likely a variable is to fail by counting the number of values in its domain, and chooses the variable with the fewest available values.
\item \textbf{DEG}: This heuristic selects the variable involved in the maximum number of constraints with unassigned variables~\cite{Bittle09}.
\item \textbf{KAPPA}: It selects the next variable such that the new subproblem minimizes the $\kappa$ factor for the whole instance~\cite{Gent96B}.
\item \textbf{WDEG}: WDEG attaches a weight to every constraint of the problem~\cite{Boussemart04}. The weights are initialized to one and increased by one whenever its respective constraint fails during the search. Then, the weighted degree of a variable is calculated as the sum of the weights of the constraints in which the variable is currently involved. WDEG gives priority to the variable with the largest weighted degree.
\end{itemize}

In all cases, the performance of the methods for solving CSPs was measured by using at least one of the following metrics~\cite{Amaya2017}:

\begin{itemize}
 \item \textbf{Consistency Checks (CC)}: Total revisions of constraints after an instance ends,
such that the larger the number of constraints, the more expensive the search becomes. This value
is used in the objective function during the training phase of the hyper-heuristics.
 \item \textbf{Adjusted Consistency Checks (ACC)}: Similar to the previous one, but discarding
instances where the solver times out.
 \item \textbf{Success Rate (SR)}: Relation between the number of completed and tested instances.
The higher the rate, the better the solver.
\end{itemize}

\section{Our Proposed Approach}
\label{sec:ProposedApproach}

We explore two fronts for improving the predictive power of features in selection hyper-heuristics: explicit 
and implicit transformations. Our motivation for doing so is twofold. The first one is that original features may change a lot throughout the first iterations, but eventually arrive at a point of negligible change. By using feature transformations this behavior could be delayed. The second one is that part of the feature space may be wasted by considering feature values that never (or scarcely) appear in practice and that belong to the same solver. Through feature transformation these regions could be compressed, raising the importance of regions that belong to different solvers. We now provide the main elements of each transformation, and throughout this 
work we also explore the eventual benefit of combining them (see Sect.~\ref{sec:Methodology}).

\subsection{Proposed Explicit Feature Transformations}
\label{sec:FeatureTransformations}

\begin{figure}
    \centering
    \includegraphics[width = 6.5cm]{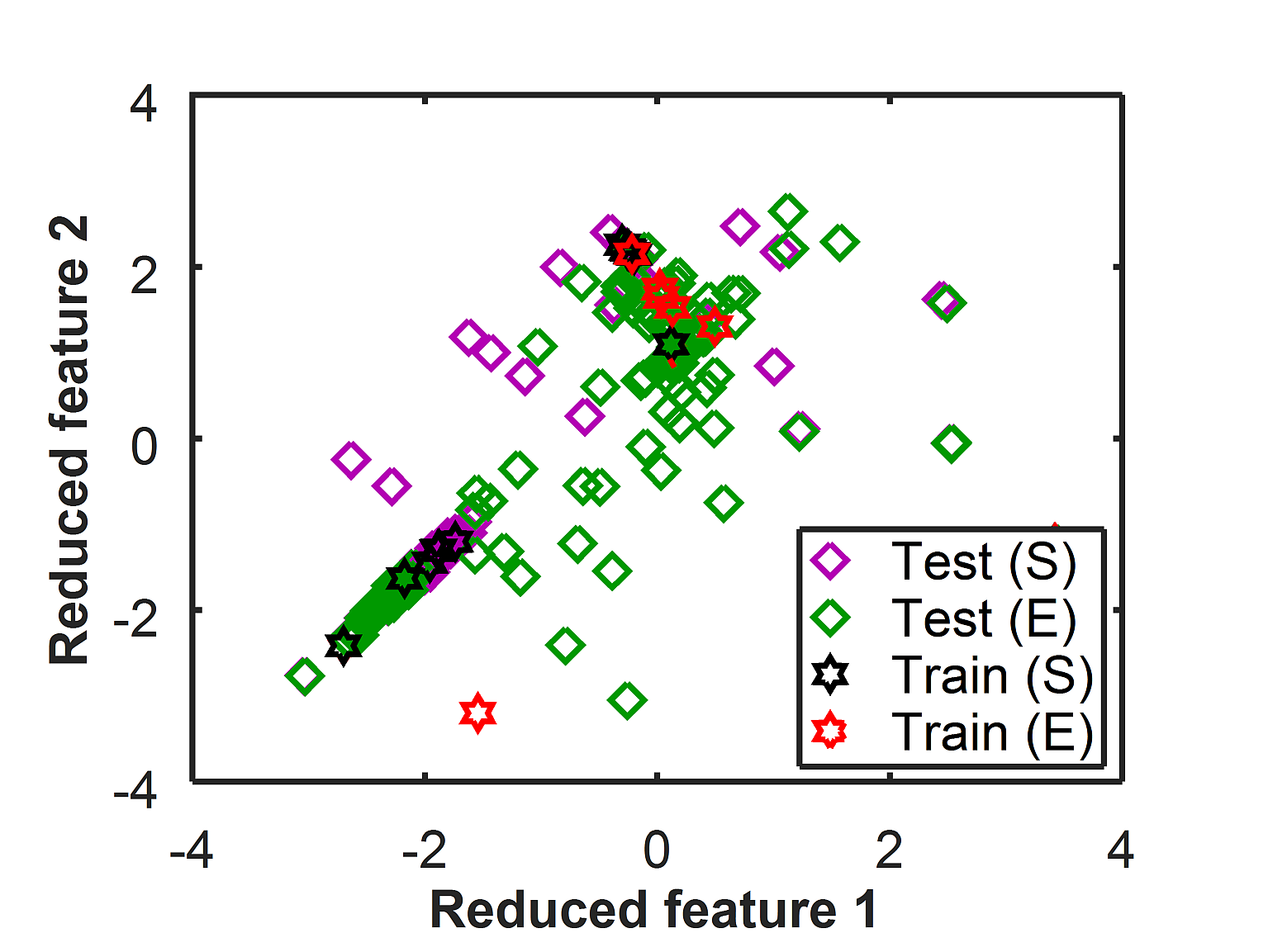}    
    \caption{Plot of initial (S) and final (E) features used in this work. Data have been reduced from eight to two features by using PCA.}
    \label{fig:FundamentalsTrainingTestingLocation}
\end{figure}

This section presents two explicit transformations for a single feature ($i$). Expressions are given in terms of a midpoint ($M_i$), and a half-width ($W_i$). For this work, we considered that every point in the training set may be meaningful within the test set and should be preserved. Thus, we defined $M_i = (\max(f_i)+\min(f_i)) / 2$ and $W_i=(\max(f_i)-\min(f_i))/2$. Here, $f_i$ is a vector containing the values of feature $i$ for the training instances.
Figure~\ref{fig:FundamentalsTrainingTestingLocation} shows the location of all instances used in this work (except for the confirmatory testing; please refer to Sect.~\ref{sec:Methodology} for more details). This plot corresponds to information yielded by a Principal Component Analysis (PCA) that was used to reduce the data from eight to two features. Moreover, data have been separated into training (stars) and testing (diamonds). Besides, the train/test ratio that will be used in the tests was also considered here. As Fig.~\ref{fig:FundamentalsTrainingTestingLocation} shows, unsolved and solved instances are spread out throughout the feature domain. Furthermore, the unsolved training instances (black stars) that seem away from the other ones actually share their location with unsolved testing instances (magenta diamonds). Therefore, if the hyper-heuristic uses this information during its training, it may perform better. Besides, as instances are solved, their features shift locations until reaching the spot indicated by red stars (training instances) and by green diamonds (testing instances). Using the transformation from~\cite{Amaya2017} does not guarantee that every value will be included. On the other hand, using the current proposal, the hyper-heuristic can adapt to the data presented in the training instances.

\begin{figure}
    \centering
    \includegraphics[width = 13cm]{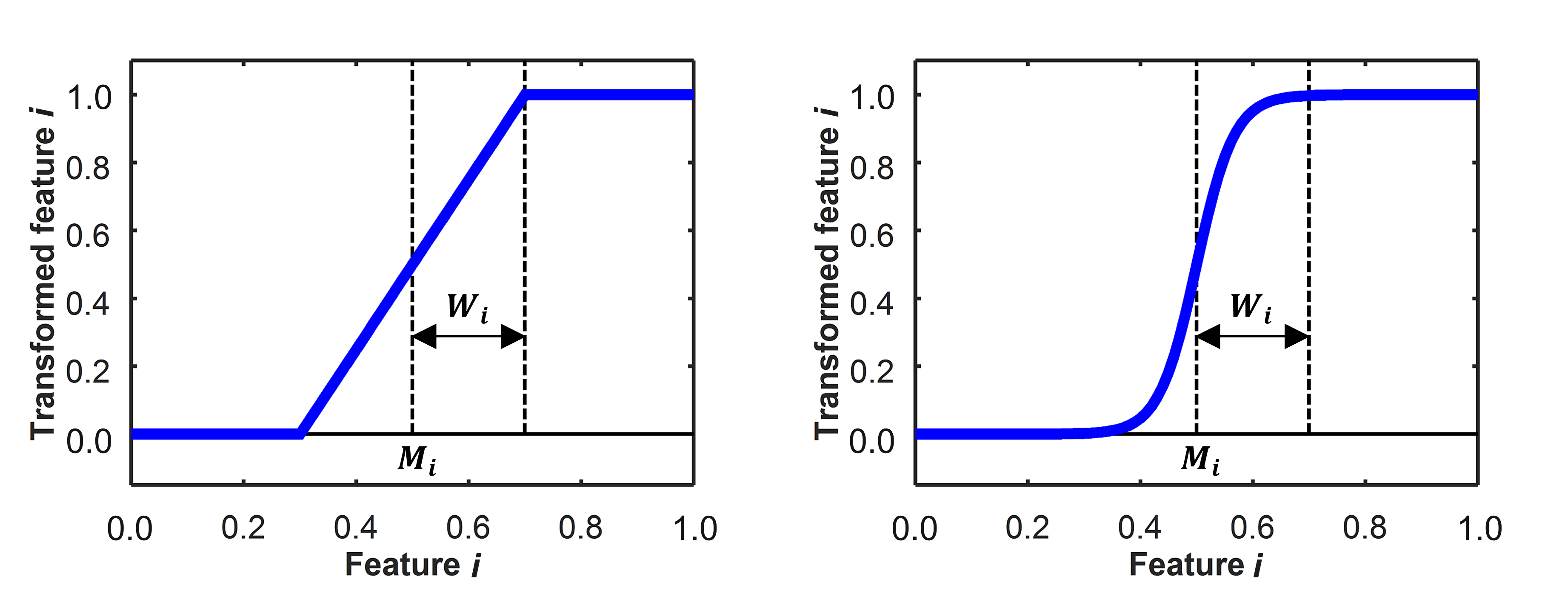}    
    \caption{Overview of the Linear (left) and S-shaped (right) transformations used in this work. $M_i$ is the center of the transformation and $W_i$ represents its half-width.}
    \label{fig:FundamentalsExplicitTransform}
\end{figure}

Figure~\ref{fig:FundamentalsExplicitTransform} shows both transformations. The idea is to map values within a given range to the full feasible interval, i.e., $[0, 1]$. In the Linear case (left), the way in which the feature is distributed remains unaffected by using Eq.~\ref{equ:LinearTransform}. In the S-shaped case (right), extreme values are smoothed out and the middle region is highlighted via Eq.~\ref{equ:ExponentialTransform}.

\begin{equation}
    \Phi_L(x_i,M_i,W_i) = \max \left(0, \min \left(1, \frac{x_i-M_i+W_i}{2W_i} \right) \right)
    \label{equ:LinearTransform}
\end{equation}    
  
\begin{equation}
    \Phi_S(x_i,M_i,W_i) = 1 - \left( \frac{ 1 }{ 1+e^{\frac{6M_i}{W_i} \left(\frac{x_i}{M_i}-1 \right) } } \right).
    \label{equ:ExponentialTransform}
\end{equation}

\subsection{Proposed Implicit Feature Transformation}
\label{sec:DistanceTransformations}

The above methods preserve low-dimensionality of the problem. This, however, can be a shortcoming when dealing with a complex distribution of instances, e.g., when instances best solved by different heuristics are very close. One way to ease this issue is by mapping to a higher dimensional space. To do so, a selection hyper-heuristic must compute the similarity between the set of rules and an instance, in this new feature space. Without loss of generality, here we assume the squared Euclidean distance. Let $\mathbf{x}^{\left(1\right)}$ and $\mathbf{x}^{\left(2\right)}$ be two points in the original $m$-dimensional feature space. Expanding the polynomial of the squared Euclidean distance reveals that, in the new feature space, this value can be computed by the inner products of the mapped instances. Thus, we use the kernel trick to perform this mapping since kernel functions allow performing an implicit mapping of features, yielding Eq.~\ref{eq:Kernel_Distance}.

\begin{equation}
d^2\left(\mathbf{x}^{\left(1\right)},\mathbf{x}^{\left(2\right)}\right) = K\left(\mathbf{x}^{\left(1\right)},\mathbf{x}^{\left(1\right)}\right)-2K\left(\mathbf{x}^{\left(1\right)},\mathbf{x}^{\left(2\right)}\right)+K\left(\mathbf{x}^{\left(2\right)},\mathbf{x}^{\left(2\right)}\right).
\label{eq:Kernel_Distance}
\end{equation}

Using kernels is a more general approach since it subsumes the previous ones. Indeed, it can be shown that the linear kernel corresponds to the original feature space. In this paper, we focus on the Radial Basis Function (RBF) kernel, because it is one of the most popular and effective in the literature~\cite{Bagnall201}. The RBF kernel, Eq.~\ref{eq:rbf_kernel}, has one adjustable parameter, which we set as $\gamma=1/N_f$, where $N_f$ is the number of features.

The RBF kernel can be regarded as a similarity measure between two instances. One way to visualize its effect is through Visual Assessment of similarity Tendency (VAT) images~\cite{bezdek2002vat}. In this kind of images, each element is compared to all others using a similarity measure. Data is usually stored in matrix form (sorted by each cluster of data). Therefore, each value in the matrix represents the similarity between the element given by the row and the one given by the column. Afterwards, an image is created to reflect each similarity value in the matrix with a color: the darker the color, the more similar elements are. Thus, black patches indicate groups of data as similar as possible, while white regions indicate maximum dissimilarity between the pair of elements.

\begin{figure}
    \centering    
    \includegraphics[width = 12.0cm]{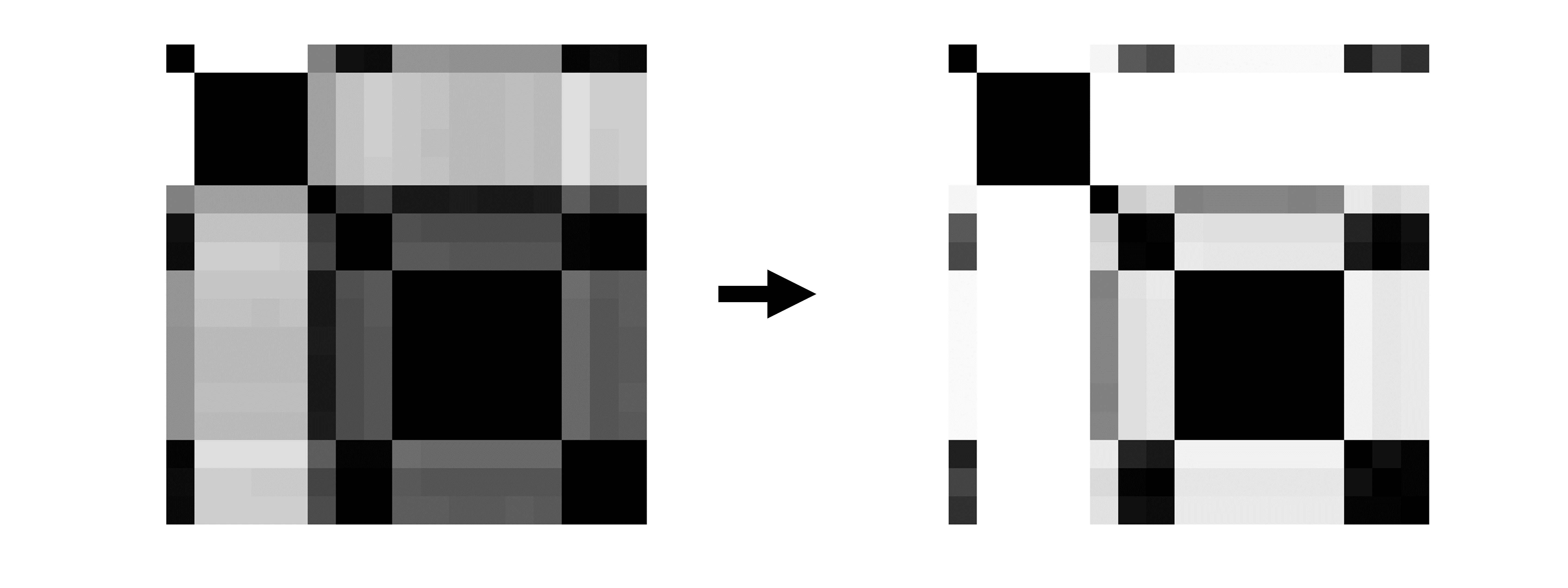}    
    \caption{VAT image using Euclidean distance (left) and kernel-based distance (right).}
    \label{fig:DistanceMatrices}
\end{figure}

\begin{figure}
    \centering
    \includegraphics{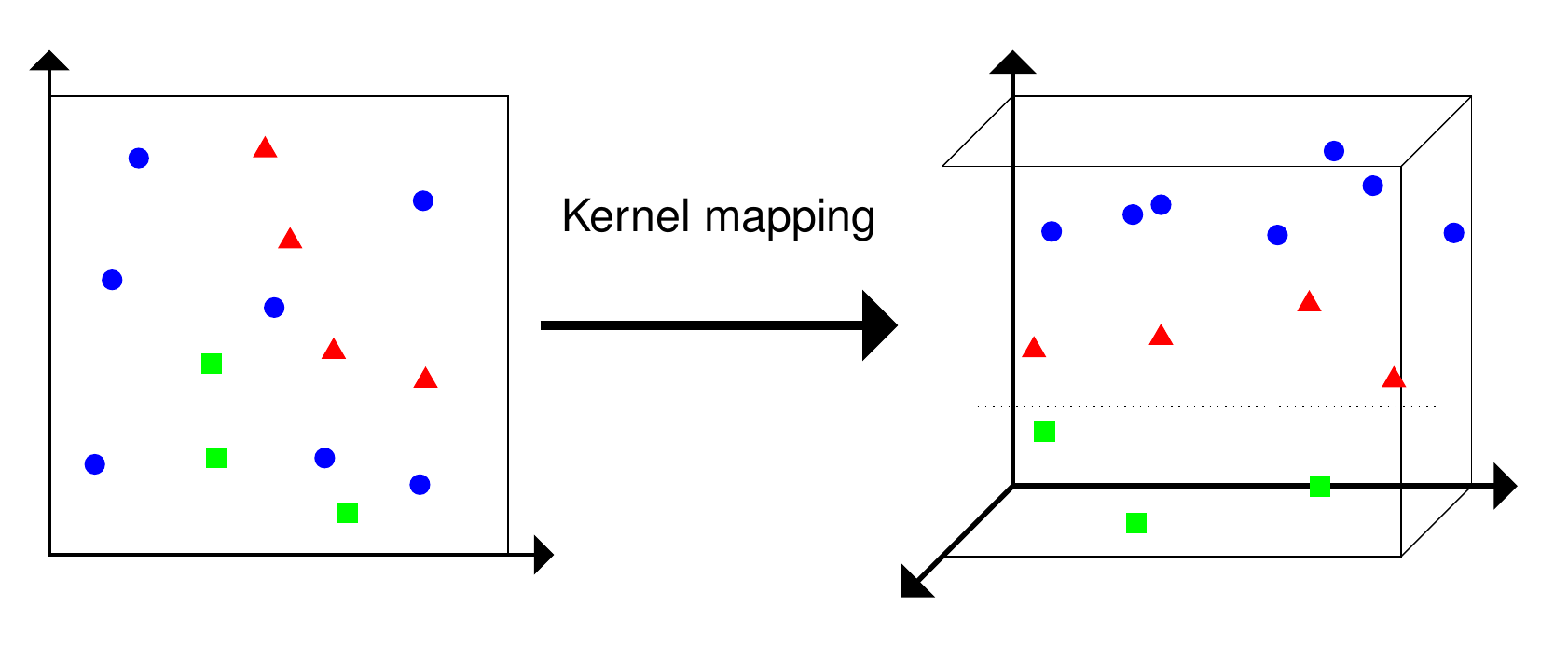}
    \caption{Separating the set of rules from a low dimensional space through a kernel mapping to a higher dimensional one.}
    \label{fig:kernel_mapping}
\end{figure}

Figure~\ref{fig:DistanceMatrices} compares a VAT image with the Euclidean distance (left) against one with the RBF kernel (right). This evidences the benefit of using a kernel (since regions are more clearly separated). Therefore, this approach swaps the traditional ``distance calculation" block from Fig.~\ref{fig:HHModel}, by one using the kernel shown in Eq.~\ref{eq:rbf_kernel}. Since selection hyper-heuristics choose a heuristic based on the similarity to a set of rules, having too many can lower their performance by creating overlapped rules. Kernel methods can overcome this issue by their implicit mapping to a higher dimensional space (Fig.~\ref{fig:kernel_mapping}).

\section{Methodology} 
\label{sec:Methodology}
\begin{figure}
    \centering
    \includegraphics[width = 13cm]{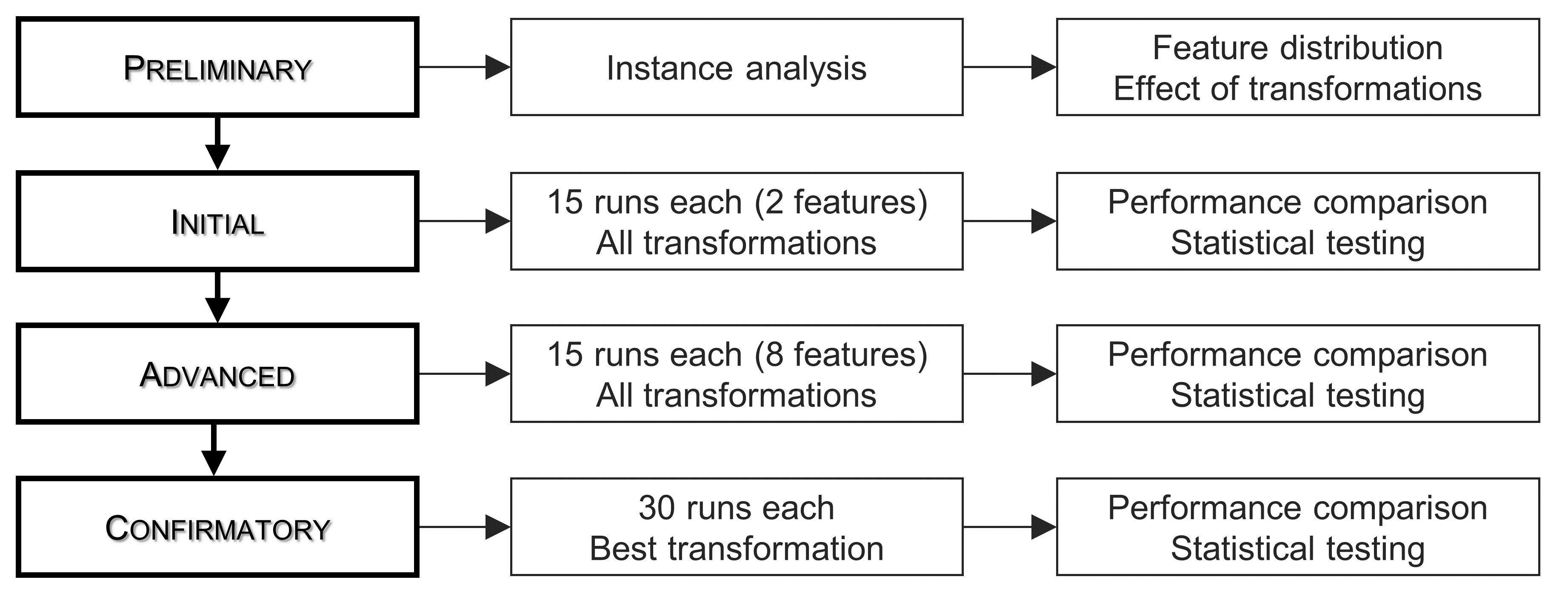}
    \caption{Overview of the four-stage methodology adopted in this work.}
    \label{fig:Methodology}
\end{figure}

We followed a four-stage methodology (Fig.~\ref{fig:Methodology}). Also, we considered the same 322 instances from~\cite{Amaya2017} split in the same way: 5\% for training and 95\% for testing. Data are publicly available\footnote{\url{https://www.cril.univ-artois.fr/~lecoutre/benchmarks.html}}, and can be identified as: \texttt{geom}, \texttt{ehi-85} and \texttt{bqwh-15-106}.

\subsection{Preliminary Testing}
This first stage tries to determine an eventual gain from using transformations to enhance the predictive power of features. Therefore, we study the distribution of original and explicitly transformed features. We also study their effect on the zones of influence of each training instance. It is worth remarking that, at this stage, the eventual gain from the kernel-based approach cannot be estimated, since its calculations are done on the fly.

\subsection{Initial Testing}
A second stage selects the features used in~\cite{Amaya2017} (i.e., $p_1$ and $p_2$). With this, we create 15 different selection hyper-heuristics for each scenario (i.e., with no transformation and for the Linear, Exponential, S-shaped, and RBF kernel). We also include experiments for the combination of kernel with Linear and S-shaped transformations. This leads to 105 selection hyper-heuristics ($15 \times 7$). First, we present their average performance. We compare it against standalone heuristics and baseline selection hyper-heuristics (i.e., with no transformation). Moreover, we compare our data against that of a synthetic oracle, which can perfectly predict the best solver for tackling each instance. Because such solver does not exist, we build it by using the best solution (among the base heuristics) for each instance. We consider it important to highlight that such a process is infeasible for a real application, but useful as a benchmark.

As a second approach, we analyze how stable the data are. We focus on the success rate and on the search cost of each selection hyper-heuristic (see Sect.~\ref{sec:CSPs}). We wrap this section up with a one-tailed Wilcoxon statistical test to determine significant increases in the performance of each approach.

\subsection{Advanced Testing}
This stage deepens the previous one by analyzing the effect of transformations over the whole set of eight features (see Sect.~\ref{sec:CSPs}). Again, we run 15 repetitions of each experiment. We also determine the performance gain of using each transformation. Moreover, we execute a one-tailed Wilcoxon statistical test to determine whether a significant performance increase can be achieved.

\subsection{Confirmatory Testing}
In this final stage, we explore the generality of our proposed approach. Therefore, we select a different domain and generate 30 base selection hyper-heuristics and 30 with the best transformation (more about this in Section~\ref{sec:ResultsConfirmatory}). In this work, we chose the knapsack problem, mainly due to its popularity and usefulness, and because knowledge about this combinatorial optimization problem is widespread. Our tests consider two sets of 600 instances each: one with 50 items, and one with 100 items. Each set was built up with the instances from~\cite{Pisinger2005}, covering groups 11 to 16. In accordance with previous tests, we trained each selection hyper-heuristic using 5\% of the instances (i.e., 30).

In this stage, selection hyper-heuristics can select among four popular heuristics. The first three select the item based on the maximum profit, the minimum weight, or the best profit/weight ratio, respectively. The fourth one selects items in their default order. Moreover, selection hyper-heuristics map an instance based on seven features calculated over the items remaining in the instance. Three of them use information from the profit, and correspond to the mean, median, and standard deviation. Another three correspond to the mean, median, and standard deviation of the weight. The final one is a measure of the correlation between profit and weight. It is important to highlight that each metric is normalized so that their values fall in the $[0,1]$ range. Therefore, the first three features are divided by the maximum profit within the instance. The next three ones are, thus, divided by the maximum weight. The final feature is increased by one and divided in half. Please bear in mind that the maximum values are dynamic as they are calculated from the items remaining in the instance.

Lastly, but not less important, it is worth mentioning that total profit is used as the metric for two events. The first one is training the selection hyper-heuristics. The second one is assessing the eventual performance gain derived from the transformation. Also, it is important to highlight that the aforementioned profit corresponds to the sum of profits achieved on each instance. As such, throughout training profit is calculated over 30 instances, but it is calculated over 570 throughout testing. As before, we run a one-tailed Wilcoxon statistical test to determine whether a significant increase in profit can be achieved.


\section{Results and Discussion} 
\label{sec:experiments}
This section presents the main results of our work. To make things easier for the reader, the structure presented in the methodology (Sect.~\ref{sec:Methodology}) is preserved, reserving one subsection for each main stage.

\subsection{Preliminary Testing} \label{sec:ResultsPreliminary}
\begin{figure}
    \centering
    \subfigure{
        \includegraphics[width = 5.2cm]{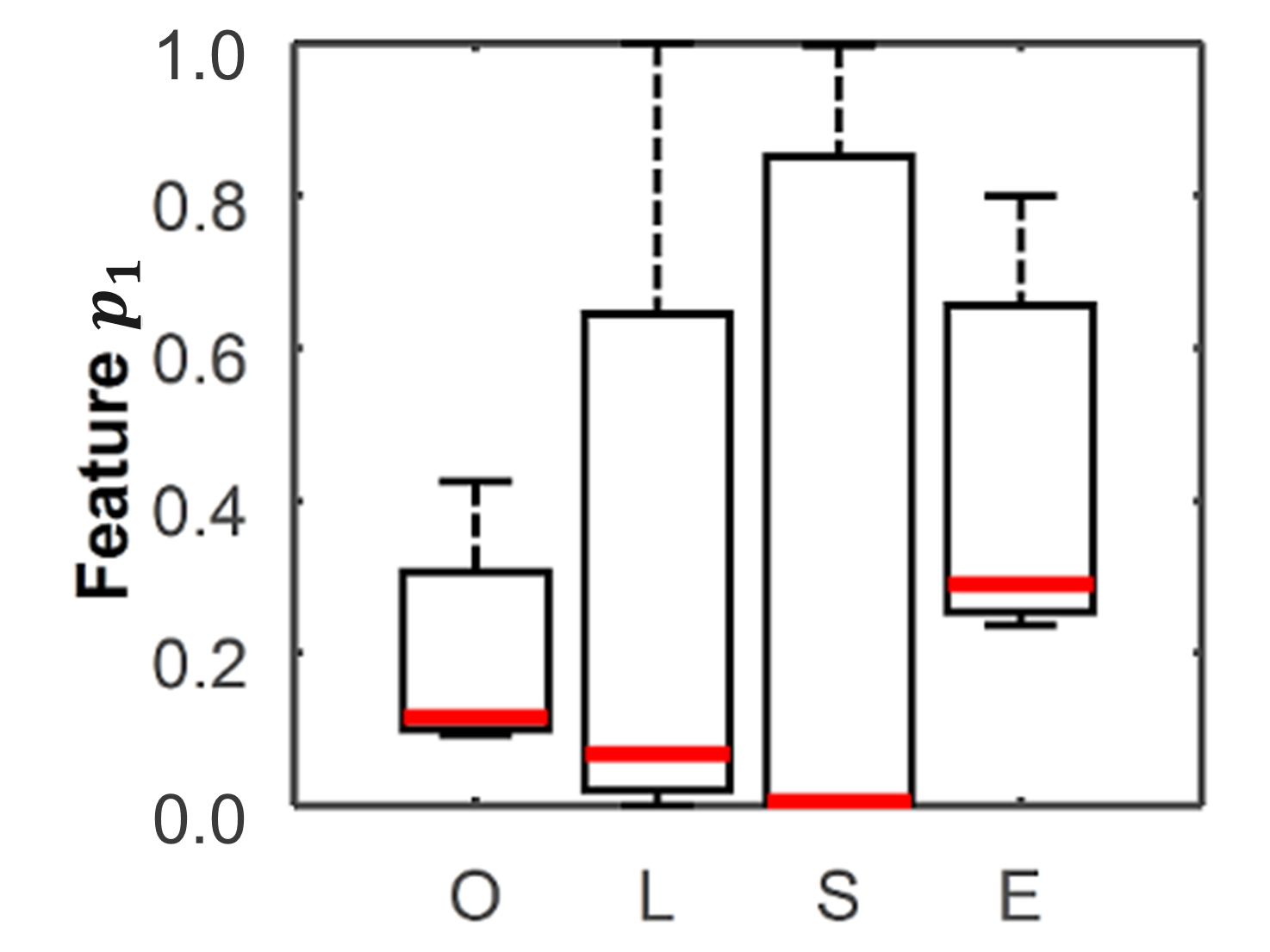}}
    \subfigure{
        \includegraphics[width = 5.2cm]{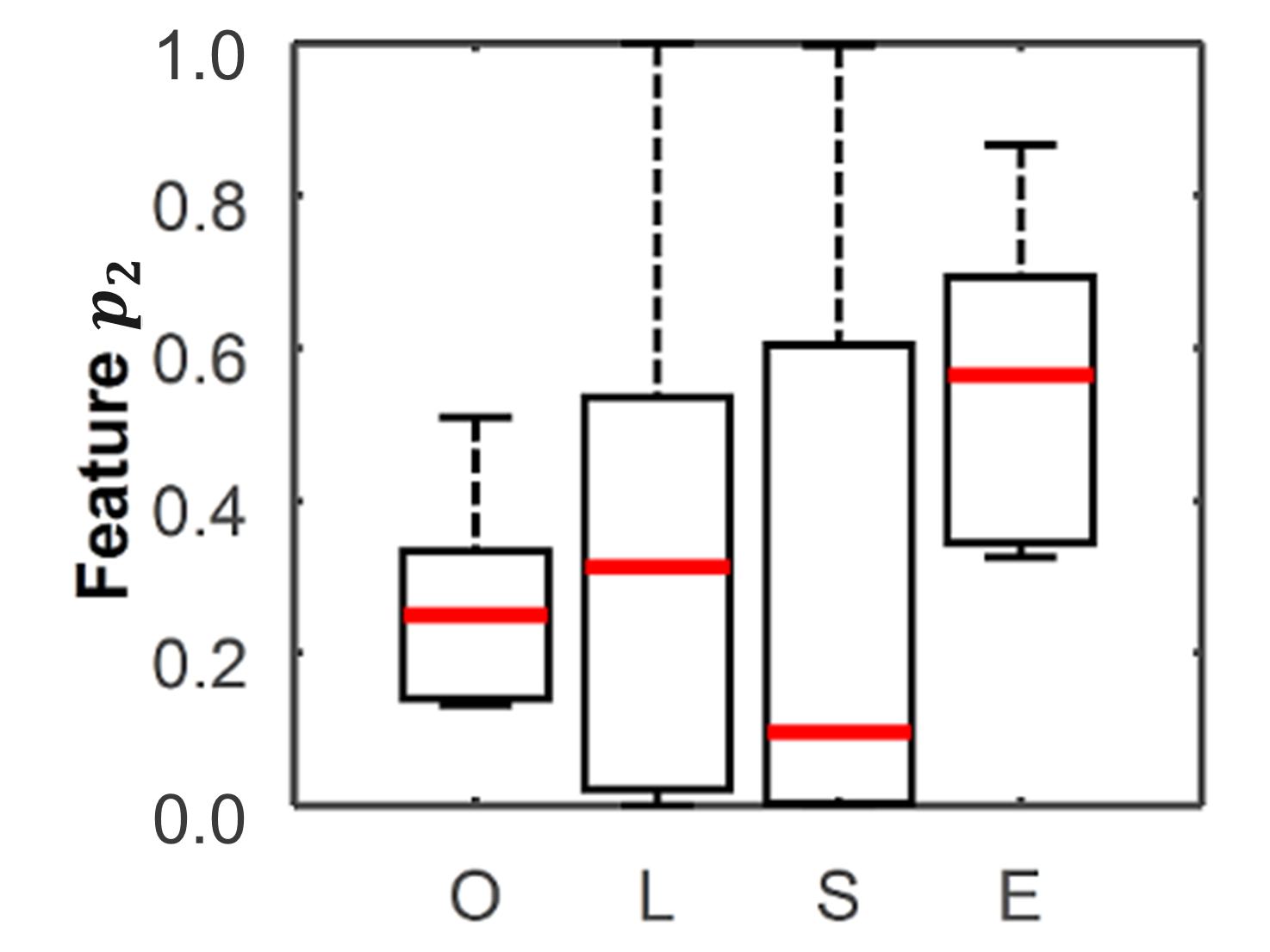}}
    \subfigure{
        \includegraphics[width = 5.2cm]{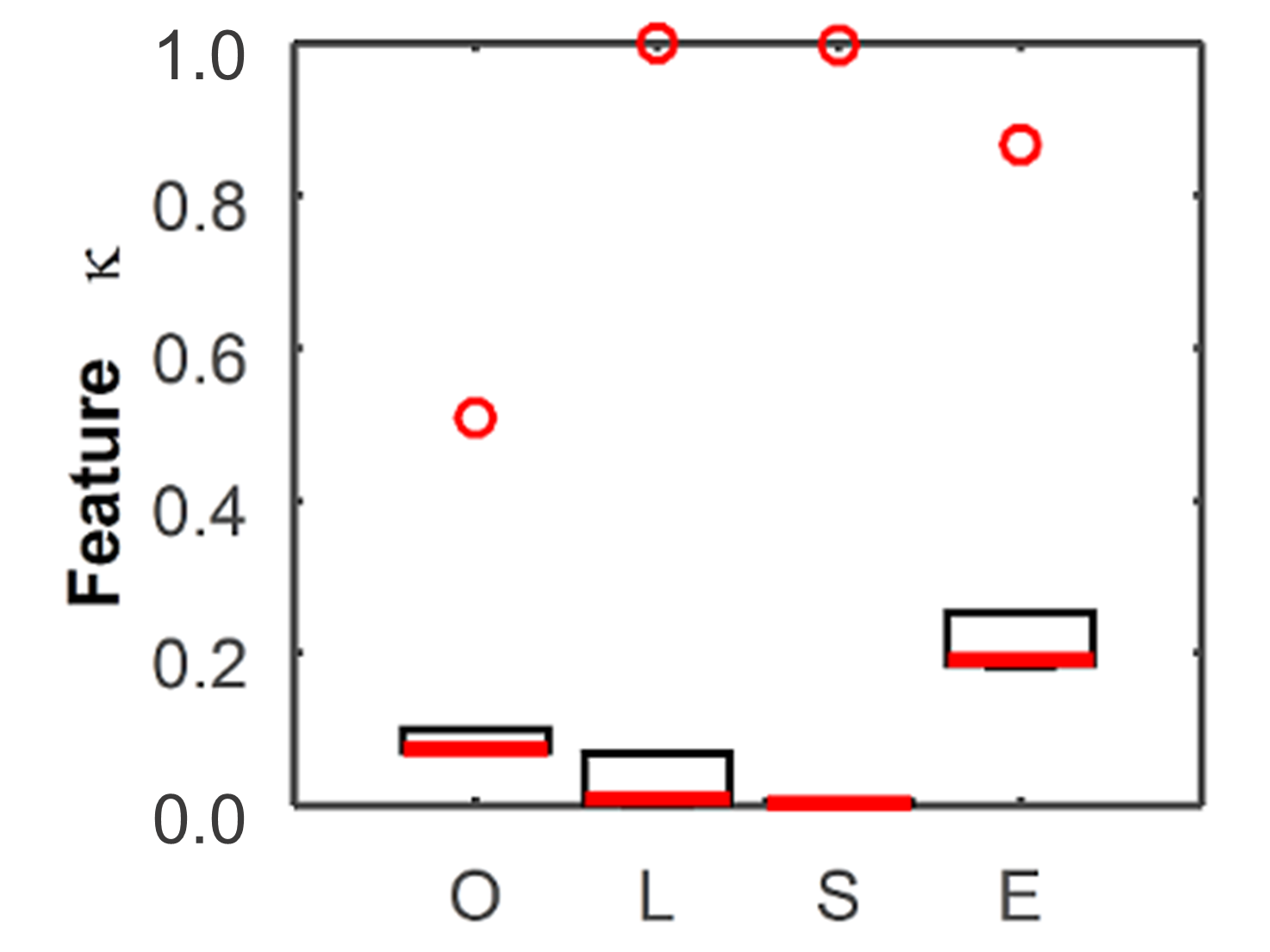}}
    \subfigure{
        \includegraphics[width = 5.2cm]{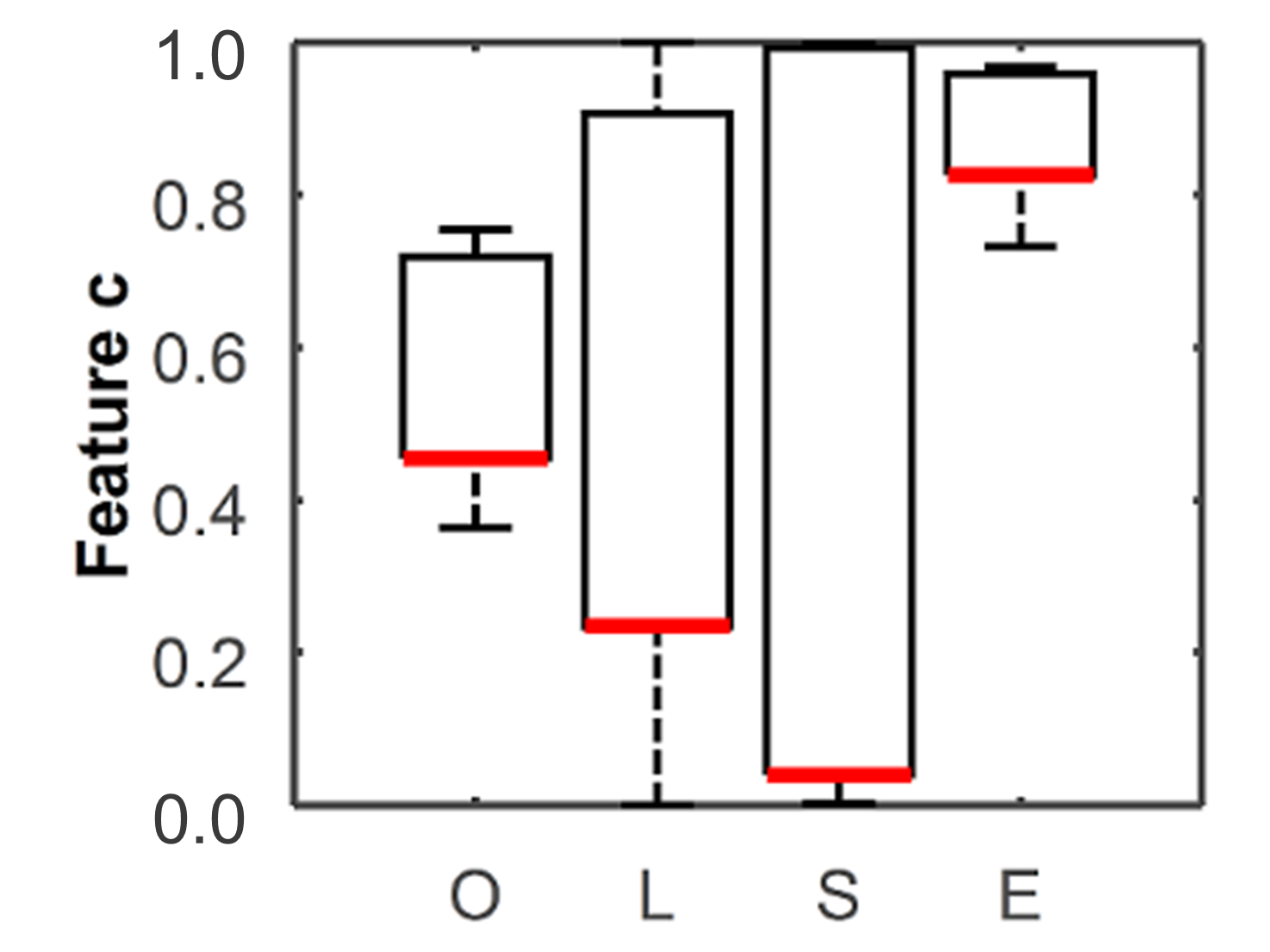}}
    \subfigure{
        \includegraphics[width = 5.2cm]{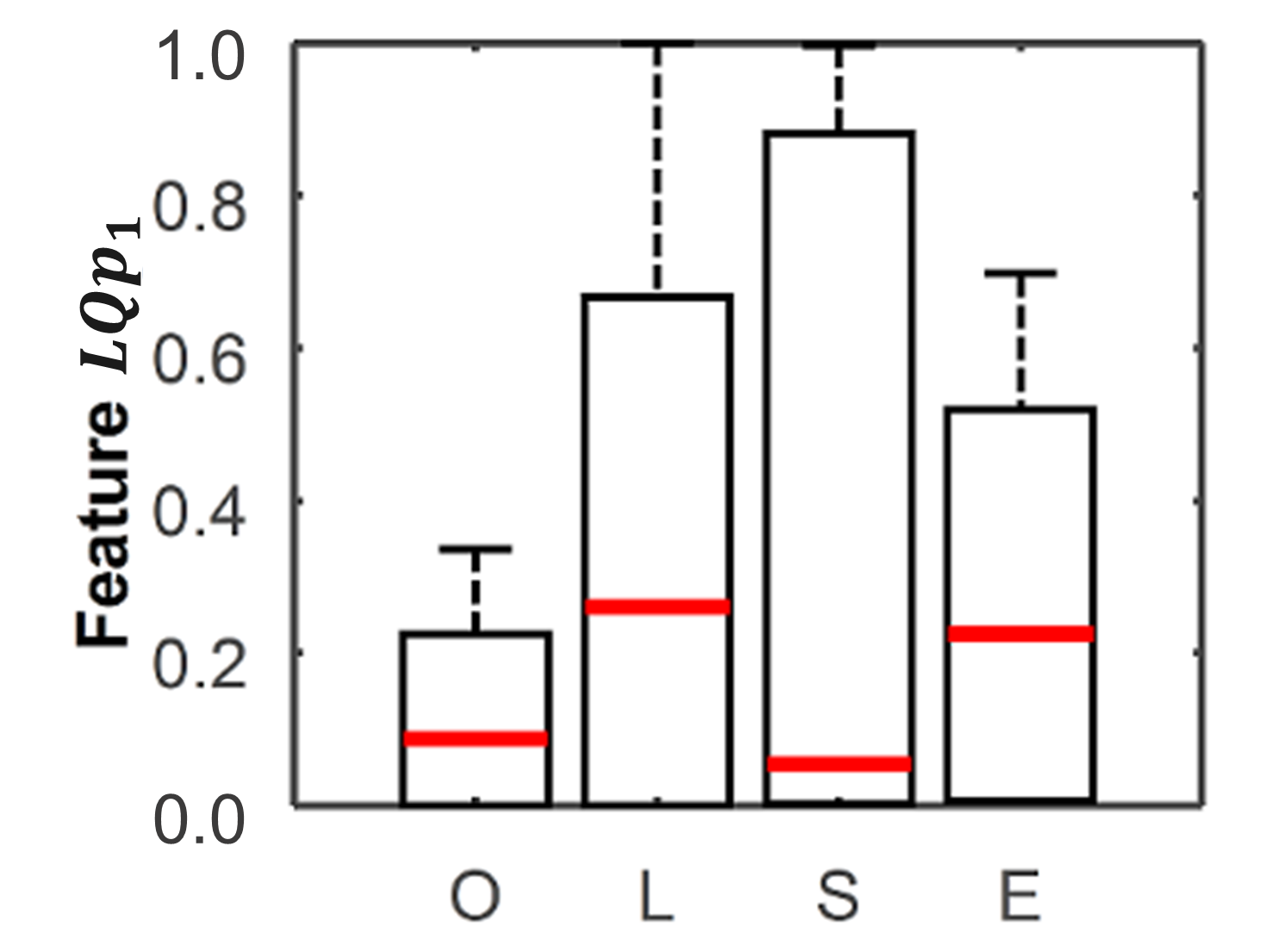}}
    \subfigure{
        \includegraphics[width = 5.2cm]{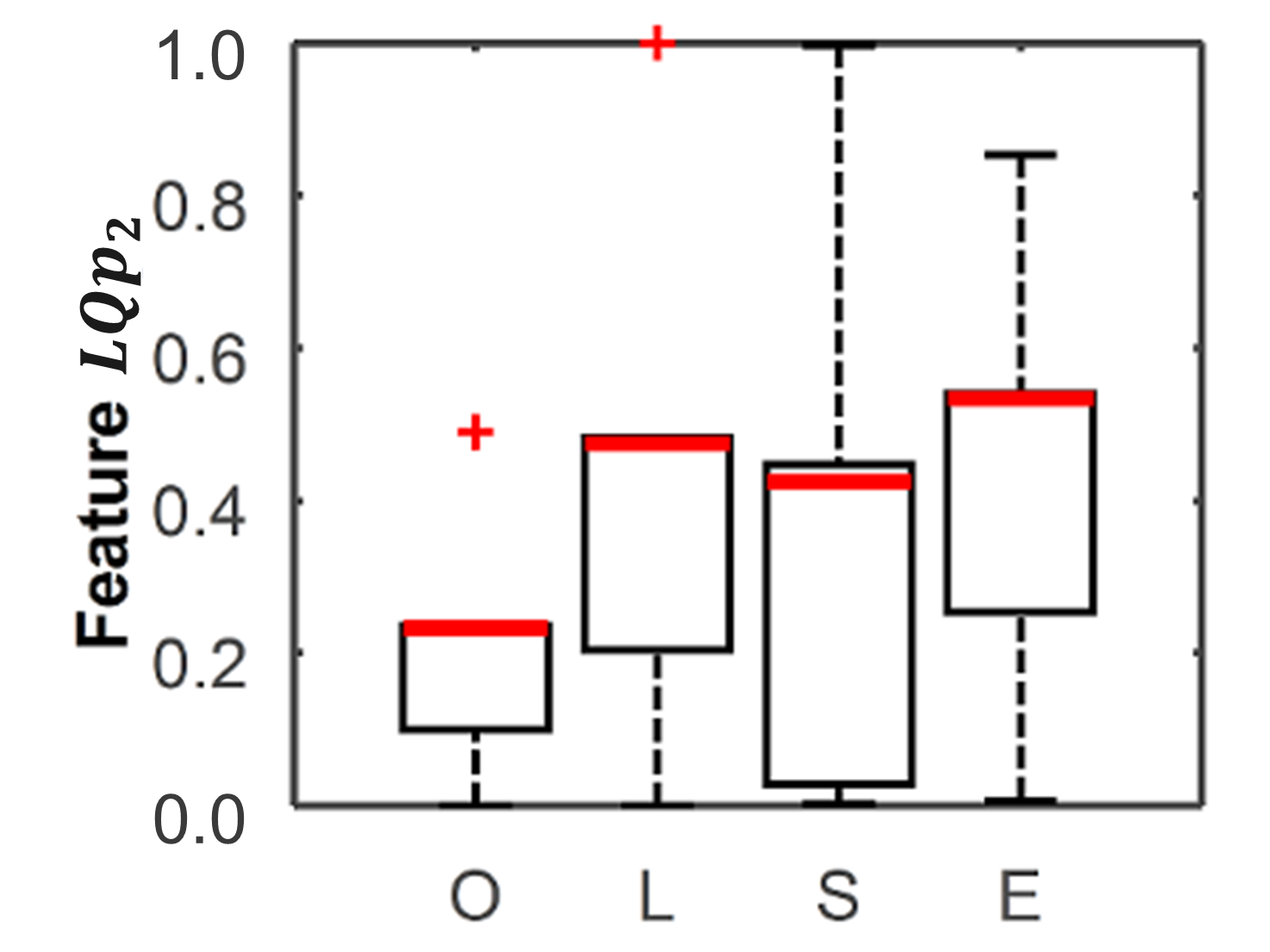}}
    \subfigure{
        \includegraphics[width = 5.2cm]{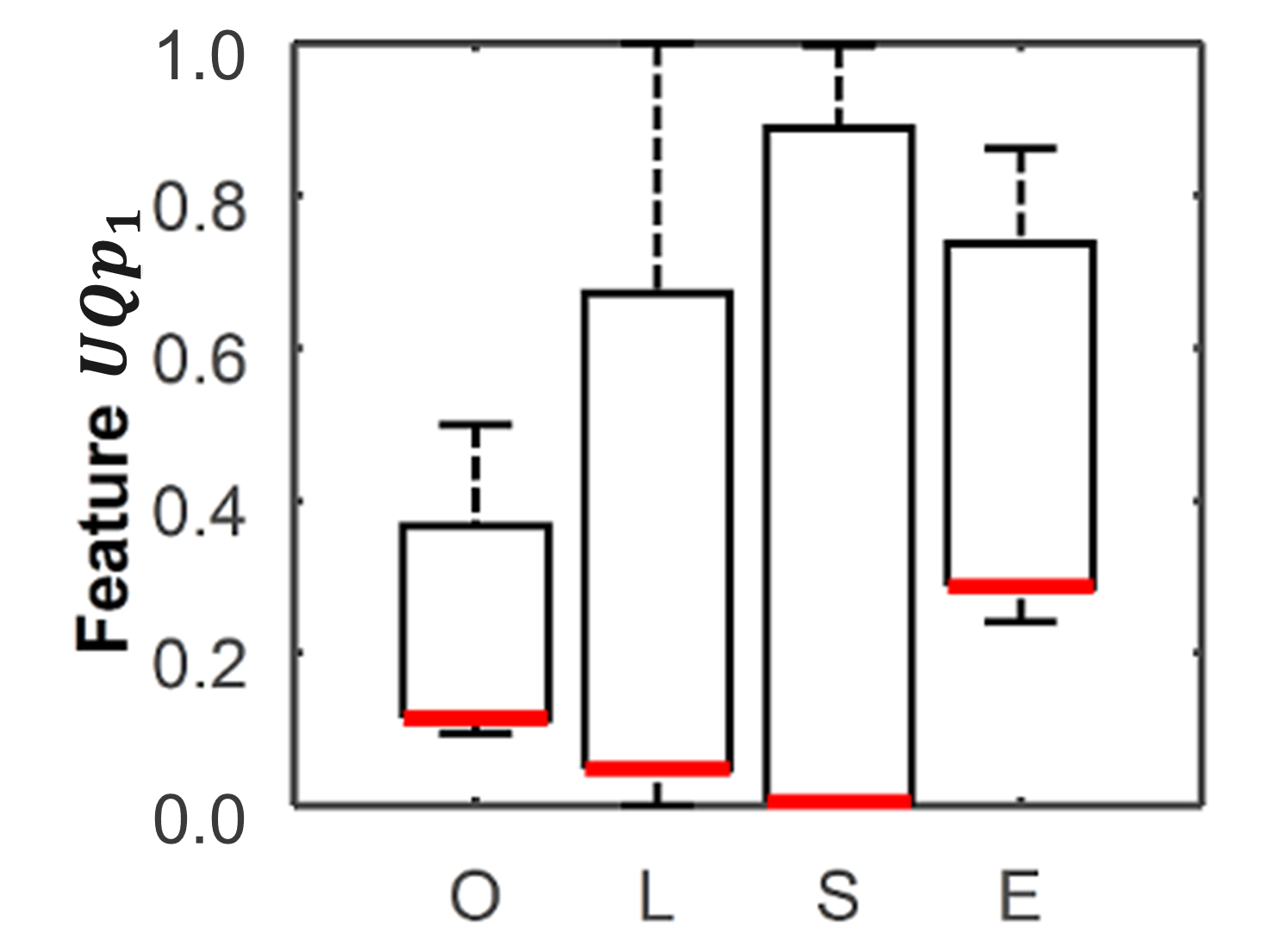}}
    \subfigure{
        \includegraphics[width = 5.2cm]{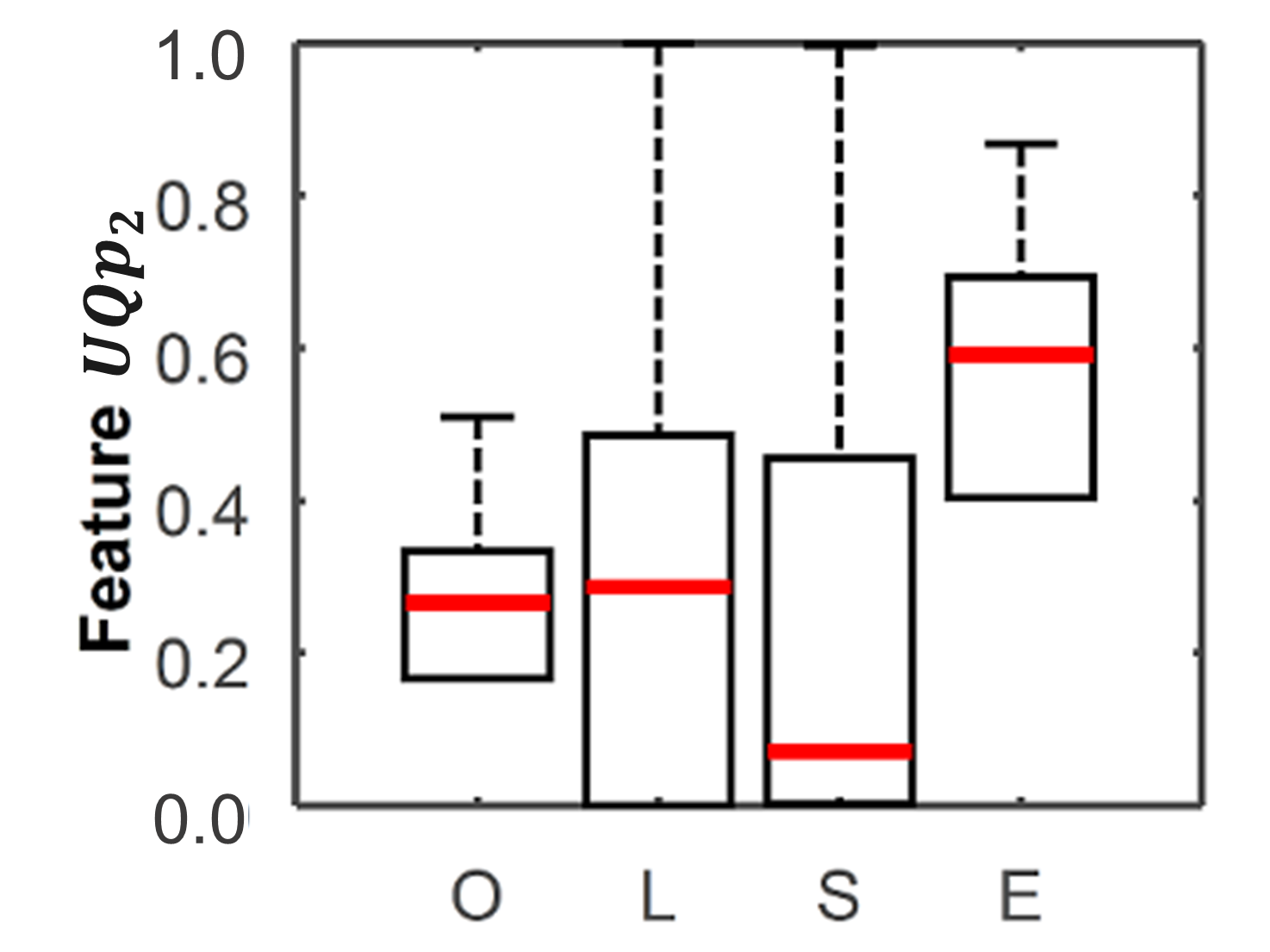}}
    \caption{Distribution of all features: Original (O) values and those with Linear (L), 
S-shaped (S), and Exponential (E) transformations. Crosses: outliers  between $1.5*IQR$ 
and $3*IQR$. Circles: outliers beyond $3*IQR$. $IQR$: Interquartile range.}
    \label{fig:ResultsPreliminaryCSPFeatures}
\end{figure}

Figure~\ref{fig:ResultsPreliminaryCSPFeatures} shows the distribution of all features in the training set, and their transformations. In most cases, the S-shaped transformation expands representative data more than the Linear transformation does. Also, in all cases the median of the former was lower than that of the latter. Another thing worth mentioning at this point is that the model used in a previous work does not expand the feature range from zero to unity. However, it allows for values all the way to zero (see Figure~\ref{fig:FundamentalsCECTransform}). This may provide it with more flexibility for advanced stages of the search where features may migrate to lower regions.

\begin{figure}
    \centering
    \subfigure{
        \includegraphics[width = 5.2cm]{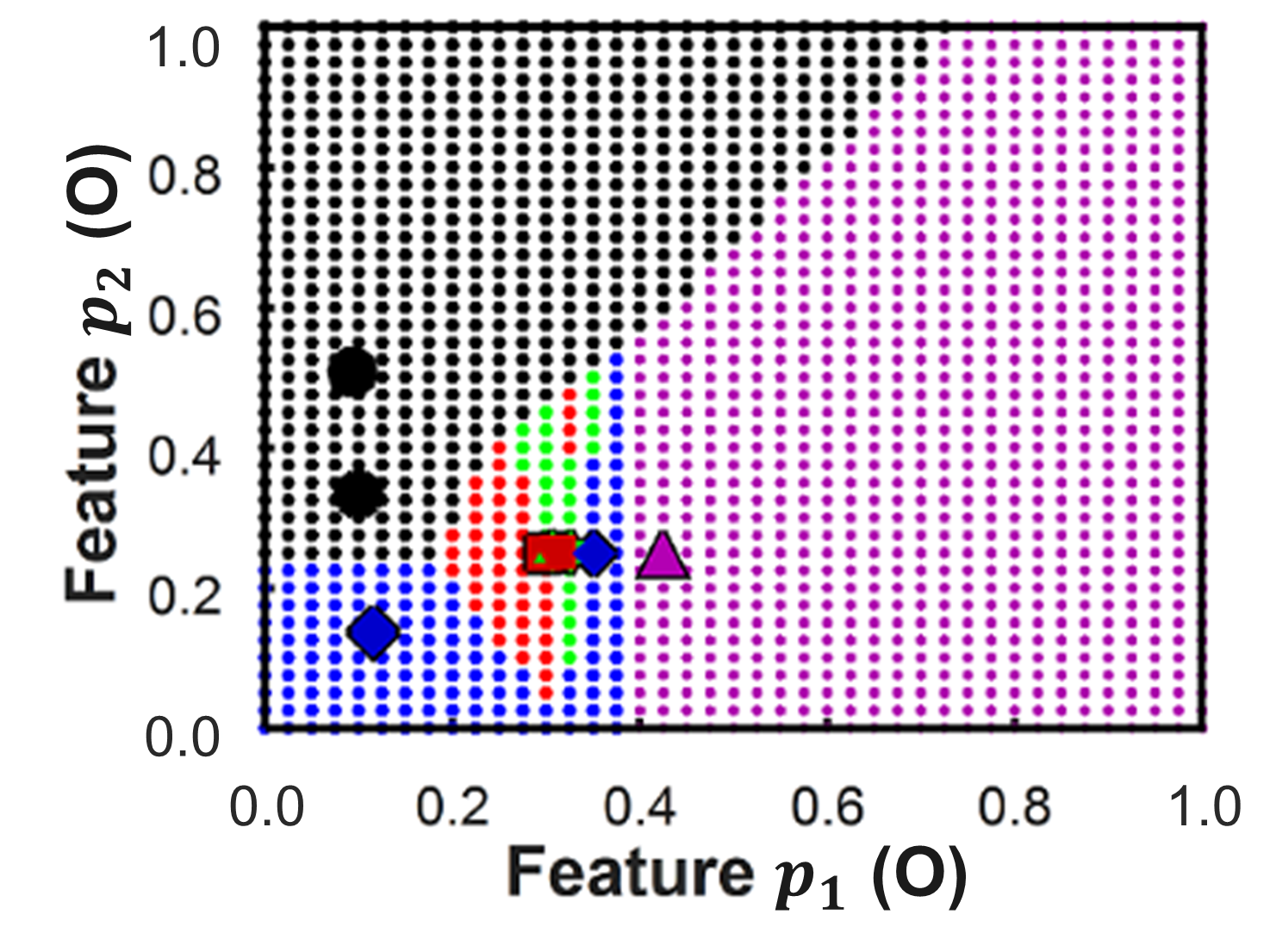}}
    \subfigure{
        \includegraphics[width = 5.2cm]{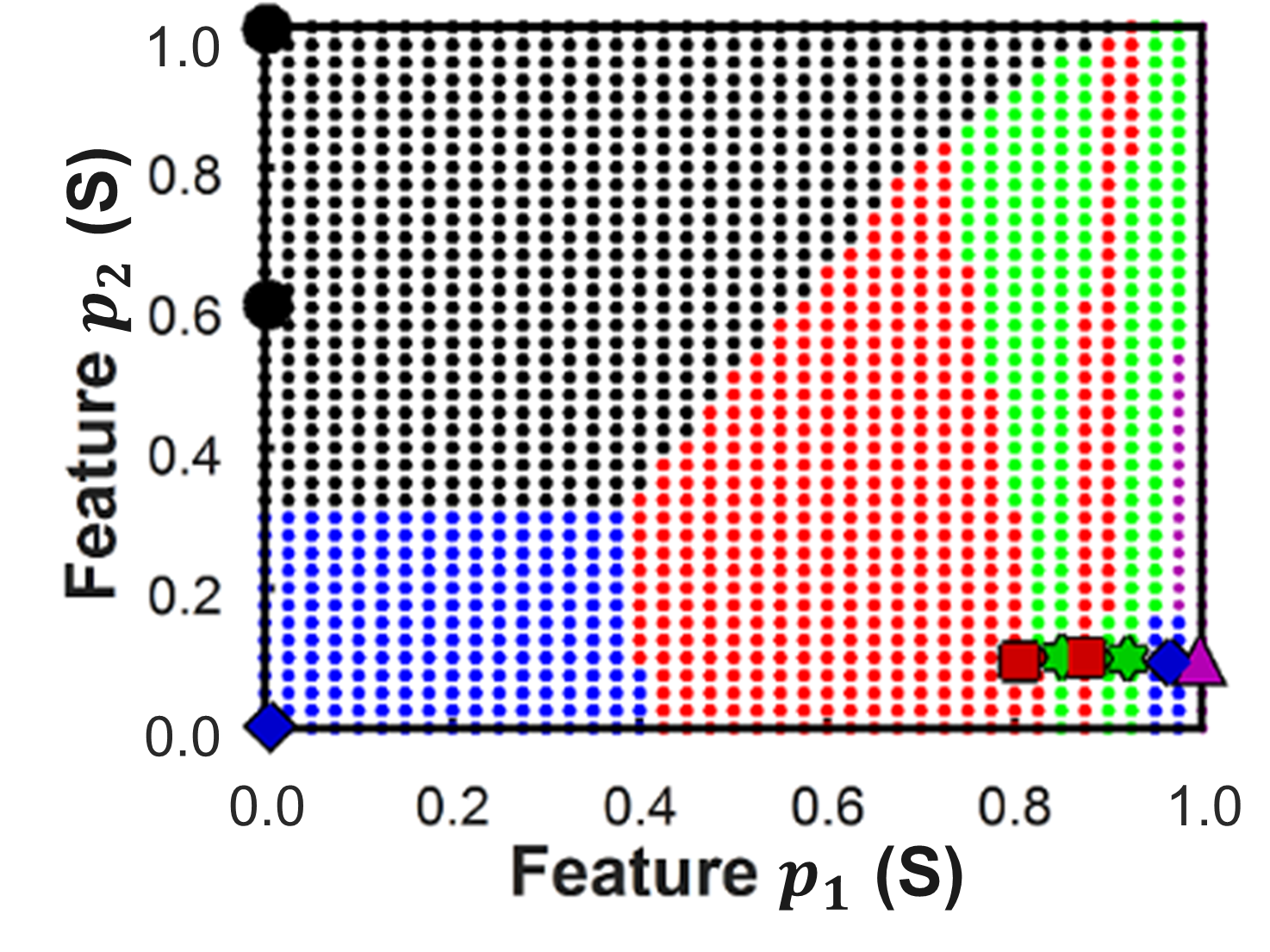}}
    \subfigure{
        \includegraphics[width = 5.2cm]{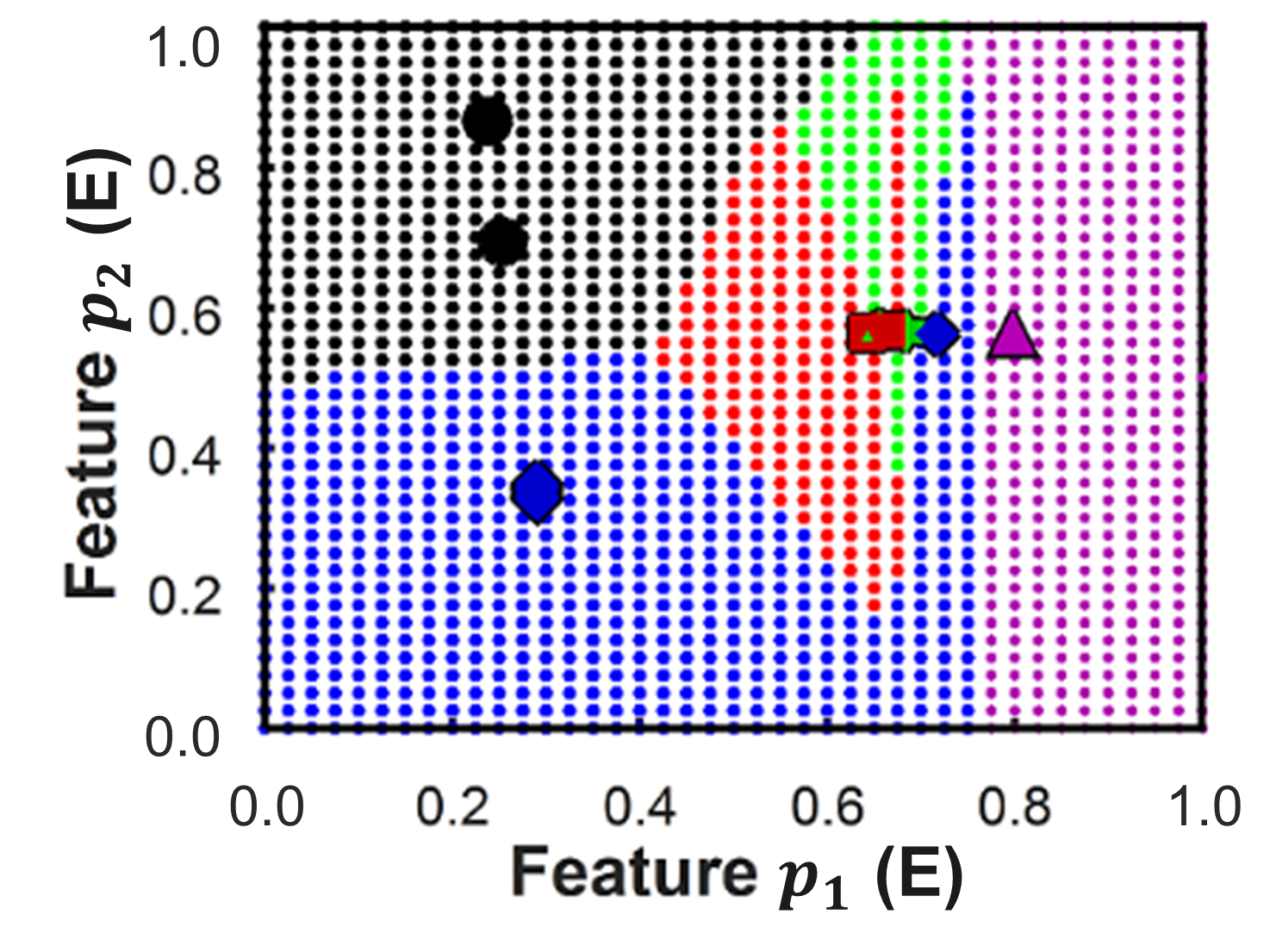}}
    \caption{Example of how transformations affect the regions of influence created by each 
instance, considering two features. Data are shown for the Original (O) values and those with 
S-shaped (S) and Exponential (E) transformations. Elements in magenta represent ties.}
    \label{fig:ResultsPreliminaryCSPZones}
\end{figure}

Figure~\ref{fig:ResultsPreliminaryCSPZones} shows the regions that each instance influences. Data are shown for the original features (left), and for the ones transformed using the S-shaped (middle) and the previously reported (right) approaches. As shown, the region beyond $p_1=0.6$ and $p_2=0.6$ is unused, and thus wasted. On the contrary, there is a region at $0.3\leq p_1\leq 0.5$ and $p_2=0.2$ (approximately) where different kinds of instances are mixed up. By using both transformations, the wasted space is reduced and distributed throughout the remaining regions, broadening the zones in conflict and thus making them easier to separate.

\subsection{Initial Testing}
\begin{figure}
    \centering    
    \includegraphics[width = 13cm]{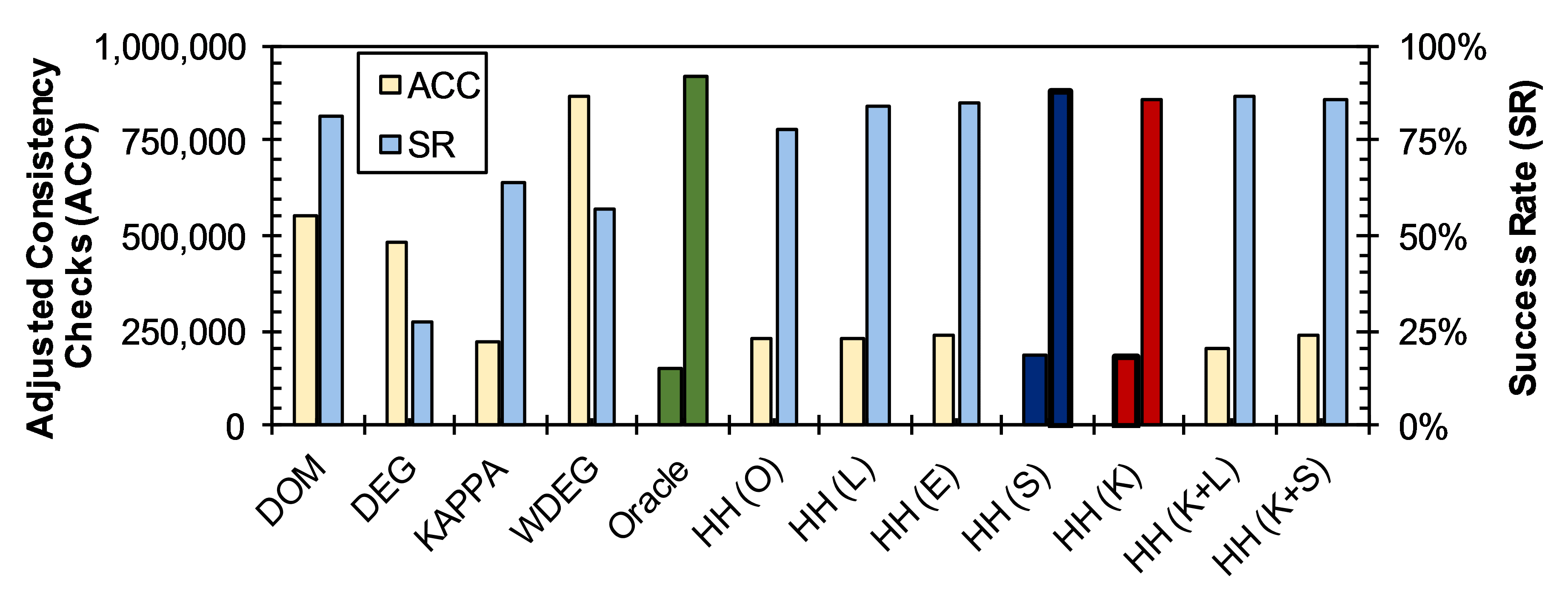}
    \caption{Average number of adjusted consistency checks (left bars) and success rate (right 
bars) for all base heuristics, for a synthetic oracle, and for all hyper-heuristics (average 
of 15 runs each) with two features: Original (O), Linear (L), Exponential (E), S-shaped (S), 
Kernel (K), Kernel+Linear (K+L), and Kernel+S-shaped (K+S). Highlighted columns: Synthetic Oracle (green), best adjusted 
consistency checks (red) and success rate (blue). Data distribution is shown in Fig.~\ref{fig:ResultsInitialPerformance}.}
    \label{fig:ResultsInitialWithHeuristics}
\end{figure}

A comparison of each base heuristic against a synthetic oracle reveals that the latter performs a lot better (Fig.~\ref{fig:ResultsInitialWithHeuristics}). Even so, this only implies that there is a latent benefit derived from an appropriate combination of each heuristic. Nonetheless, Oracle data were generated in a synthetic fashion by analyzing the performance of each standalone heuristic at each instance and selecting the best one. Thus, it represents a Utopian scenario where a perfect selection was carried out. In spite of this, and as it was expected, all selection hyper-heuristics (including those with no transformation) performed better than standalone heuristics. Moreover, all transformations exhibited an average performance quite close to that of the synthetic oracle (highlighted bar in green) in terms of both, number of adjusted consistency checks and success rate. Even so, the S-shaped transformation completed, on average, a bit more instances than the other approaches (highlighted bar in blue). The kernel-based approach was computationally cheapest (highlighted bar in red). It is also important to remark that, even the worst transformations were not so bad. In fact, the Linear transformation yielded a success rate 3\% higher than the best performing heuristic (i.e., DOM) while requiring about 60\% less ACC. Besides, the combination of kernel with S-shaped transformations required only 7\% more ACC than the best heuristic (i.e., KAPPA) but increased the success rate in 22\%. A comparison of the average behavior of selection hyper-heuristics with no transformation is also interesting. For this, we focus on search cost and on success rate. The best transformations shifted the success rate by 8\% and by 10\% (respectively), while requiring about 20\% less ACC (in both cases).

\begin{figure}
    \centering
    \subfigure{
        \includegraphics[width = 6.5cm]{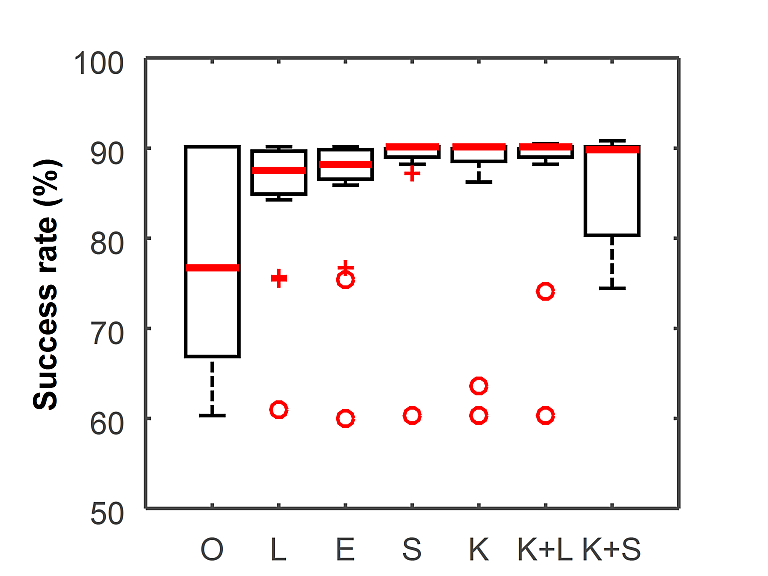}}
    \subfigure{
        \includegraphics[width = 6.5cm]{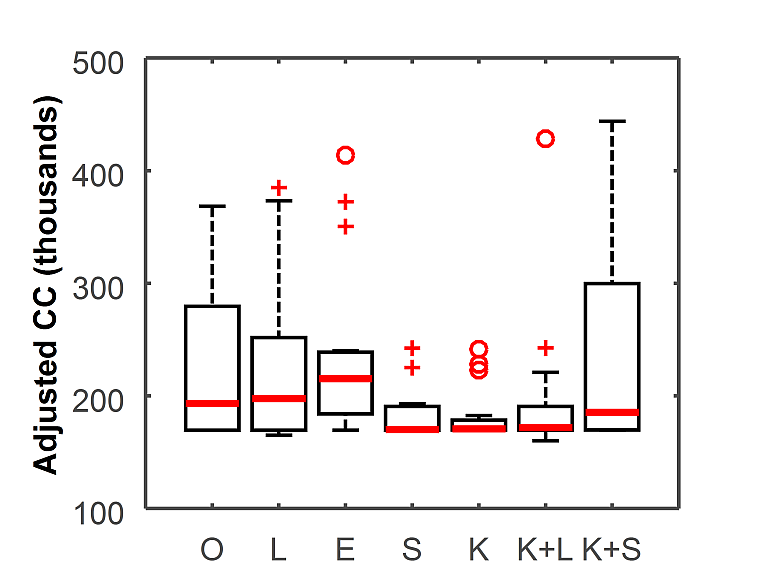}}    
    \caption{Success rate (left) and search cost (right) for 15 runs of hyper-heuristics operating with 
two features. Original values (O) and transformations: Linear (L), S-shaped (S), Exponential (E), Kernel 
(K), Kernel+Linear (K+L), and Kernel+S-shaped (K+S). Crosses: outliers between $1.5*IQR$ and $3*IQR$. Circles: outliers beyond $3*IQR$. $IQR$: Interquartile range.}
    \label{fig:ResultsInitialPerformance}
\end{figure}

Selection hyper-heuristics were distributed in an interesting fashion (see Fig.~\ref{fig:ResultsInitialPerformance}). For starters, all transformations increased the median success rate in over 10\%. Two of them (S-shaped and kernel) also reduced the median cost of the search in about 20,000 consistency checks. Alas, the Linear and the previously proposed transformations (identified as `Exponential') led to a more computationally expensive search path, increasing the median number of consistency checks. However, this number corresponds to the number of validations that must be performed in those instances where the solver could find a solution within the time limit. Therefore, this increase in cost could be derived from the additional instances that were solved. Figure~\ref{fig:ResultsInitialPerformance} also includes data for the combinations of explicit and implicit transformations. We did so striving to analyze whether merging them led to a better performance. As shown, even if the performance improves, only the Linear transformation yields a behavior similar to that of the kernel. Though seemingly enhancing the success rate, it hinders the search cost. For the S-shaped transformation, however, including the kernel-based distance hampers performance, leading to less successful and more costly selection hyper-heuristics.

A Wilcoxon statistical test yielded the p-values shown in Table~\ref{tab:Pvalue2F}. The S-shaped and both mixed (i.e., K+L and K+S) transformations had a higher success rate than the original approach (p-values below 0.05). Even so, the p-value of the pure kernel (0.0548) was not too high, which makes it an alternative worth keeping in mind. Analyzing the search cost reveals that no approach boasted a huge improvement over the original one. Nonetheless, the S-shaped transformation performed best, with a p-value of 0.0855.

\begin{table}[ht!]
\renewcommand{\familydefault}{\sfdefault}\normalfont
\centering
\caption{P-values of the Wilcoxon statistical test (two features), between the original approach and each modification. L: Linear. E: Exponential. S: S-shaped. K: Kernel. K+L: Kernel+Linear. K+S: Kernel+S-shaped. SR: Success rate. ACC: Adjusted consistency checks.}
\label{tab:Pvalue2F}
\begin{tabular}{@{}lcccccc@{}}
\toprule
\textbf{Metric} & \textbf{L} & \textbf{E} & \textbf{S} & \textbf{K} & \textbf{K+L} & \textbf{K+S} \\ \midrule
\textbf{SR}     & 0.2107     & 0.2502     & 0.0325     & 0.0548     & 0.0343       & 0.0492       \\
\textbf{ACC}    & 0.4098     & 0.7466     & 0.0855     & 0.2214     & 0.1313       & 0.5659       \\ \bottomrule
\end{tabular}
\end{table}

\subsection{Advanced Testing}

\begin{figure}
    \centering
    \subfigure{
        \includegraphics[width = 6.0cm]{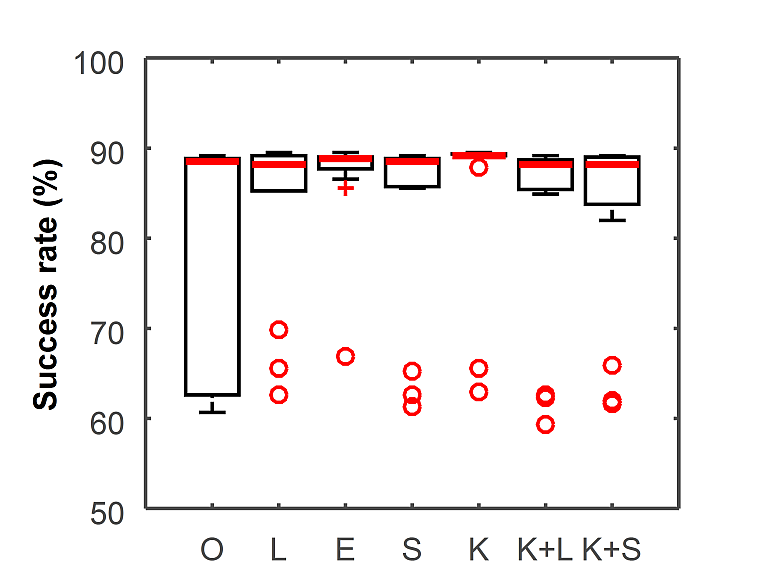}}
    \subfigure{
        \includegraphics[width = 6.0cm]{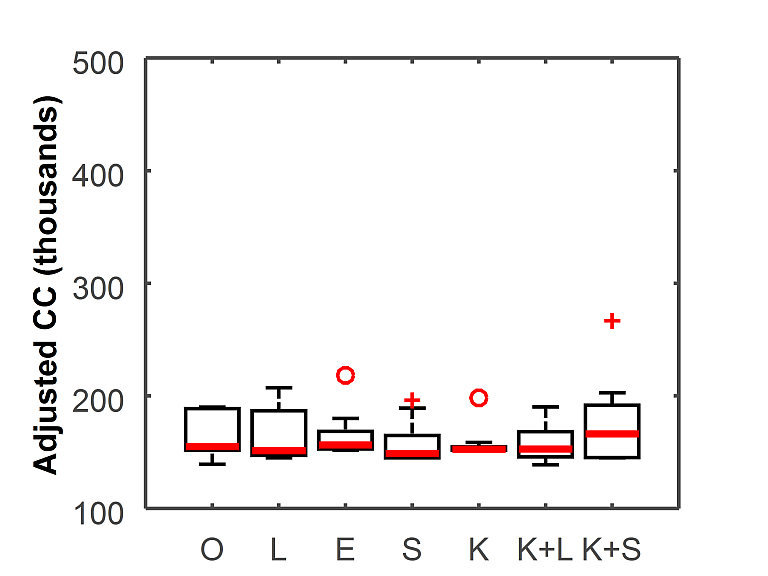}}    
    \caption{SR (left) and ACC (right) for 15 runs of selection hyper-heuristics operating 
with eight features. Original values (O) and transformations: Linear (L), S-shaped (S), Exponential (C), 
Kernel (K), Kernel+Linear (K+L), and Kernel+S-shaped (K+S). Crosses: outliers between $1.5*IQR$ and $3*IQR$. 
Circles: outliers beyond $3*IQR$. $IQR$: Interquartile range.}
    \label{fig:ResultsAdvancedPerformance}
\end{figure}

Migrating to more features worked in favor of selection hyper-heuristics (Fig.~\ref{fig:ResultsAdvancedPerformance}). The median SR of all solvers increased to about 90\% (even that of the selection hyper-heuristic with no transformation). Nonetheless, all transformations were still helpful since they increased the stability of this metric (i.e., SR), yielding only a few outliers with poor performance. The median number of required consistency checks was also reduced in about 50,000. Moreover, it was reduced to about one half for the worst performing hyper-heuristic. Even so, and similarly than with two features, the kernel-based approach was the one which reduced the most the variation in the cost of the search. Furthermore, again, combining both kinds of approach yielded mixed results. In fact, it was helpful for the Linear transformation. It is, however, not so much for the S-shaped one, since they became less successful and more costly than without the kernel.

This time around, the Wilcoxon statistical test revealed the p-values shown in Table~\ref{tab:Pvalue8F}. As can be seen, the pure kernel approach performed remarkably well (p-value of 0.0003). This makes it virtually safe to assume that its success rate is significantly higher than that of the original approach. Another result worth mentioning is that of the exponential transformation (p-value of 0.0276). Despite this, no approach provided a p-value below 0.10 when testing the search cost. Nevertheless, the pure kernel approach exhibited a standard deviation reduced by almost 40\%.

\begin{table}[ht!]
\renewcommand{\familydefault}{\sfdefault}\normalfont
\centering
\caption{P-values of the Wilcoxon statistical test (eight features), between the original approach and each modification. L: Linear. E: Exponential. S: S-shaped. K: Kernel. K+L: Kernel+Linear. K+S: Kernel+S-shaped. SR: Success rate. ACC: Adjusted consistency checks.}
\label{tab:Pvalue8F}
\begin{tabular}{@{}lcccccc@{}}
\toprule
\textbf{Metric} & \textbf{L} & \textbf{E} & \textbf{S} & \textbf{K} & \textbf{K+L} & \textbf{K+S} \\ \midrule
\textbf{SR}     & 0.1577     & 0.0276     & 0.3145     & 0.0003     & 0.5335       & 0.2858       \\
\textbf{ACC}    & 0.2668     & 0.5000     & 0.1497     & 0.1403     & 0.1595       & 0.5659       \\ \bottomrule
\end{tabular}
\end{table}

\subsection{Confirmatory Testing} \label{sec:ResultsConfirmatory}
\begin{figure}
    \centering
    \subfigure{
        \includegraphics[width = 6.5cm]{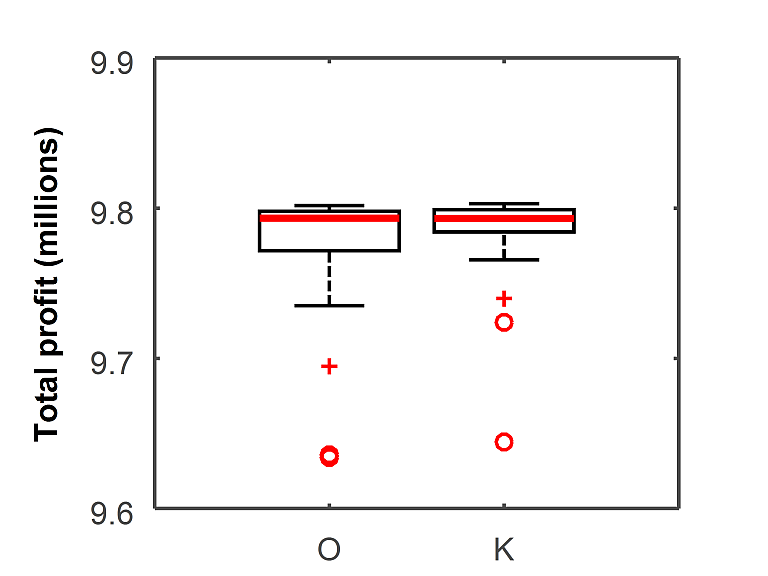}}
    \subfigure{
        \includegraphics[width = 6.5cm]{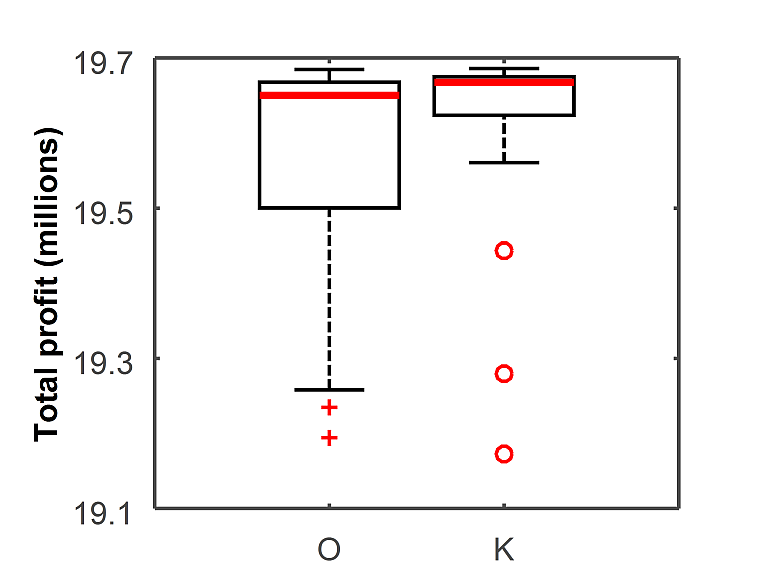}}
    \caption{Total profit distribution for the original (O) and kernel-based (K) selection hyper-heuristics (30 runs each). Left: instances with 50 items. Right: instances with 100 items.}
    \label{fig:ResultsConfirmatoryKnapsack}
\end{figure}

The pure kernel transformation was the only one that proved to be statistically better than the original approach during the previous phases. Hence, we selected it as our best approach and used it for this stage. Figure~\ref{fig:ResultsConfirmatoryKnapsack} shows the distribution of the performance achieved by 30 selection hyper-heuristics with no transformation (O), and by 30 with the kernel-based distance (K). The standard deviation of the profit was reduced by almost 30\% (going from 44,275 to 31,900), for instances with 50 items (left). For instances with 100 items (right), it was reduced by almost 20\% (going from 143,175 to 118,864). Moreover, statistical evidence (p-value of 0.02073) supports the claim that kernel-based distance produces, on average, more competent heuristics than Euclidean distance. Our data suggests that, as the problem increases in difficulty (represented by more items), the approach that relies on the kernel-based distance remains more stable.

\section{Conclusions and Future Work} 
\label{sec:conclusions}
Throughout this work we defined two approaches for carrying out an explicit transformation of features, and one for doing so implicitly. Our aim was to improve the performance of selection hyper-heuristics. We tested these ideas, and their combinations, on the widely used domain of CSPs. We found that, when considering two features, all transformations help the selection hyper-heuristic to generate solvers that perform closer (on average) to a synthetic oracle. Moreover, the S-shaped and the kernel-based transformations increased the median SR in about 15\% while decreasing the median search cost in about 20,000 ACC. This means that both approaches led to a higher number of instances being solved while reducing the cost of solving them. Even so, combining both ideas did not prove fruitful. Linear transformation was the only one enhanced by adding the kernel, improving its success rate but hindering its search cost. 

Increasing the number of features up to eight did not work as well for the S-shaped transformation as it did for the kernel. This time around the median success rate of both approaches lurked around 90\%, but only the latter exhibited almost no variation (except for a few outliers). Similarly, the search cost was better through both approaches. In fact, the median search cost of the S-shaped transformation was the lowest one, being 5,000 adjusted consistency checks lower than that without the transformation. Only the one using kernel had a small variation, exhibiting a standard deviation almost 40\% lower than that of the original approach.

Because of the aforementioned, we determined that the pure Kernel approach was the best way of incorporating transformations into selection hyper-heuristics. A confirmatory test run on the Knapsack domain considered instances with 50 and 100 items. Data revealed that the standard deviation of the profit achieved by hyper-heuristics could be reduced by almost 30\% and 20\% (respectively), making transformations a worthwhile effort. Moreover, a statistical test confirmed a significant increase in the performance of selection hyper-heuristics for the set with 100 items. Thus, we recommend following this idea and applying it to different domains and under different conditions, as to better assess how its benefits propagate.

In this work, we defined $\gamma$ as the inverse of the number of features. However, some exploratory tests (omitted due to space restrictions) revealed that this may not always be the best approach for selecting it. Therefore, a future research avenue in this path could relate to improving the kernel-based transformation. This could be done by designing a procedure that tailors $\gamma$ to each problem domain. In other words, one that tailors it to different configurations, e.g., by using information not only from the number of features but also from their nature. Another path worth following is to carry out a more extensive testing, e.g., by increasing the number and variety of instances and features. Such testing should include a careful analysis of all the different combinations of features, focusing on the effect of feature transformations on the performance of hyper-heuristics.

\section*{Acknowledgement}

This research was partially supported by CONACyT Basic Science Projects under grants 241461, 
221551 and 287479, and ITESM Research Group with Strategic Focus in Intelligent Systems.


\begin{thebibliography}{10}
\providecommand{\url}[1]{#1}
\csname url@samestyle\endcsname
\providecommand{\newblock}{\relax}
\providecommand{\bibinfo}[2]{#2}
\providecommand{\BIBentrySTDinterwordspacing}{\spaceskip=0pt\relax}
\providecommand{\BIBentryALTinterwordstretchfactor}{4}
\providecommand{\BIBentryALTinterwordspacing}{\spaceskip=\fontdimen2\font plus
\BIBentryALTinterwordstretchfactor\fontdimen3\font minus
  \fontdimen4\font\relax}
\providecommand{\BIBforeignlanguage}[2]{{%
\expandafter\ifx\csname l@#1\endcsname\relax
\typeout{** WARNING: IEEEtran.bst: No hyphenation pattern has been}%
\typeout{** loaded for the language `#1'. Using the pattern for}%
\typeout{** the default language instead.}%
\else
\language=\csname l@#1\endcsname
\fi
#2}}
\providecommand{\BIBdecl}{\relax}
\BIBdecl

\bibitem{Burke13}
E.~K. Burke, M.~Gendreau, M.~Hyde, G.~Kendall, G.~Ochoa, E.~{\"O}zcan, and
  R.~Qu, ``Hyper-heuristics: A survey of the state of the art,'' \emph{Journal
  of the Operational Research Society}, vol.~64, no.~12, pp. 1695--1724, Dec.
  2013.

\bibitem{ortiz2013learning}
J.~C. Ortiz-Bayliss, H.~Terashima-Mar{\'\i}n, and S.~E. Conant-Pablos,
  ``Learning vector quantization for variable ordering in constraint
  satisfaction problems,'' \emph{Pattern Recognition Letters}, vol.~34, no.~4,
  pp. 423--432, Mar. 2013.

\bibitem{Ferreira2017}
T.~N. Ferreira, J.~A.~P. Lima, A.~Strickler, J.~N. Kuk, S.~R. Vergilio, and
  A.~Pozo, ``Hyper-heuristic based product selection for software product line
  testing,'' \emph{IEEE Computational Intelligence Magazine}, vol.~12, no.~2,
  pp. 34--45, Apr. 2017.

\bibitem{guyon2003introduction}
I.~Guyon and A.~Elisseeff, ``An introduction to variable and feature
  selection,'' \emph{Journal of Machine Learning Research}, vol.~3, no. Mar,
  pp. 1157--1182, Mar. 2003.

\bibitem{garcia2014data}
S.~Garc{\'\i}a, J.~Luengo, and F.~Herrera, \emph{Data Preprocessing in Data
  Mining}.\hskip 1em plus 0.5em minus 0.4em\relax Springer, 2014, vol.~72.

\bibitem{Xue2016}
B.~Xue and M.~Zhang, ``{Evolutionary computation for feature manipulation: Key
  challenges and future directions},'' in \emph{2016 IEEE Congress on
  Evolutionary Computation (CEC)}.\hskip 1em plus 0.5em minus 0.4em\relax
  Vancouver, Canada: IEEE, Jul. 2016, pp. 3061--3067.

\bibitem{Xue2016a}
B.~Xue, M.~Zhang, W.~N. Browne, and X.~Yao, ``{A survey on evolutionary
  computation approaches to feature selection},'' \emph{IEEE Transactions on
  Evolutionary Computation}, vol.~20, no.~4, pp. 606--626, Aug. 2016.

\bibitem{Pyle1999Book}
D.~Pyle, \emph{Data Preparation for Data Mining}, 1st~ed.\hskip 1em plus 0.5em
  minus 0.4em\relax San Francisco, CA, USA: Morgan Kaufmann Publishers Inc.,
  1999.

\bibitem{smith2012measuring}
K.~Smith-Miles and L.~Lopes, ``Measuring instance difficulty for combinatorial
  optimization problems,'' \emph{Computers \& Operations Research}, vol.~39,
  no.~5, pp. 875--889, May 2012.

\bibitem{Montazeri2016}
M.~Montazeri, ``{HHFS: Hyper-heuristic feature selection},'' \emph{Intelligent
  Data Analysis}, vol.~20, no.~4, pp. 953--974, Jun. 2016.

\bibitem{Hart2017}
E.~Hart, K.~Sims, K.~Kamimura, and B.~Gardiner, ``{A hybrid method for feature
  construction and selection to improve wind-damage prediction in the forestry
  sector},'' in \emph{GECCO '17 Proceedings of the Genetic and Evolutionary
  Computation Conference}, Berlin, Germany, Jul. 2017, pp. 1121--1128.

\bibitem{Amaya2017}
I.~Amaya, J.~C. Ortiz-Bayliss, A.~E. Guti{\'{e}}rrez-Rodr{\'{i}}guez,
  H.~Terashima-Mar{\'{i}}n, and C.~A. {Coello Coello}, ``{Improving
  hyper-heuristic performance through feature transformation},'' in \emph{2017
  IEEE Congress on Evolutionary Computation}, San Sebastian, Spain, Jun. 2017,
  pp. 2614--2621.

\bibitem{scholkopf2002learning}
B.~Sch{\"o}lkopf and A.~J. Smola, \emph{Learning with Kernels: Support Vector
  Machines, Regularization, Optimization, and Beyond}.\hskip 1em plus 0.5em
  minus 0.4em\relax MIT press, 2002.

\bibitem{Wolpert1997}
D.~H. Wolpert and W.~G. Macready, ``No free lunch theorems for optimization,''
  \emph{IEEE Transactions on Evolutionary Computation}, vol.~1, no.~1, pp.
  67--82, Apr. 1997.

\bibitem{Rice76}
J.~R. Rice, ``The algorithm selection problem,'' \emph{Advances in Computers},
  vol.~15, pp. 65--118, 1976.

\bibitem{OMahony2008}
E.~O'Mahony, E.~Hebrard, A.~Holland, C.~Nugent, and B.~O'Sullivan, ``Using
  case-based reasoning in an algorithm portfolio for constraint solving,'' in
  \emph{Irish Conference on Artificial Intelligence and Cognitive Science},
  Cork City, Ireland, Aug. 2008, pp. 210--216.

\bibitem{Malitsky2012}
Y.~Malitsky and M.~Sellmann, ``Instance-specific algorithm configuration as a
  method for non-model-based portfolio generation,'' in \emph{Integration of AI
  and OR Techniques in Constraint Programming for Combinatorial Optimization
  Problems. CPAIOR 2012. Lecture Notes in Computer Science}, N.~Beldiceanu,
  N.~Jussien, and {\'E}.~Pinson, Eds.\hskip 1em plus 0.5em minus 0.4em\relax
  Berlin, Heidelberg: Springer Berlin Heidelberg, 2012, pp. 244--259.

\bibitem{Ortiz2016}
J.~C. Ortiz-Bayliss, H.~Terashima-Mar{\'i}n, and S.~E. Conant-Pablos, ``Combine
  and conquer: An evolutionary hyper-heuristic approach for solving constraint
  satisfaction problems,'' \emph{Artificial Intelligence Review}, vol.~46,
  no.~3, pp. 327--349, Oct. 2016.

\bibitem{Marin2006}
J.~G.~Mar\'{i}n-Bl\'{a}zquez and S.~Schulenburg, ``Multi-step environment
  learning classifier systems applied to hyper-heuristics,'' in \emph{GECCO '06
  Proceedings of the 8th Annual Conference on Genetic and Evolutionary
  Computation}, Seattle, USA, Jul. 2006, pp. 1521--1528.

\bibitem{Ortiz13C}
J.~C. Ortiz-Bayliss, H.~Terashima-Mar\'{i}n, and S.~E. Conant-Pablos, ``Using
  learning classifier systems to design selective hyper-heuristics for
  constraint satisfaction problems,'' in \emph{2013 IEEE Congress on
  Evolutionary Computation (CEC 2013)}, Cancun, Mexico, Jun. 2013, pp.
  2618--2625.

\bibitem{Vapnik1995Book}
V.~N. Vapnik, \emph{The Nature of Statistical Learning Theory}.\hskip 1em plus
  0.5em minus 0.4em\relax New York, NY, USA: Springer-Verlag New York, Inc.,
  1995.

\bibitem{Berlier10}
J.~Berlier and J.~McCollum, ``A constraint satisfaction algorithm for
  microcontroller selection and pin assignment,'' in \emph{Proceedings of the
  IEEE SoutheastCon 2010 (SoutheastCon)}, Concord, USA, Mar. 2010, pp.
  348--351.

\bibitem{Bochkarev15}
S.~V. Bochkarev, M.~V. Ovsyannikov, A.~B. Petrochenkov, and S.~A. Bukhanov,
  ``Structural synthesis of complex electrotechnical equipment on the basis of
  the constraint satisfaction method,'' \emph{Russian Electrical Engineering},
  vol.~86, no.~6, pp. 362--366, Jun. 2015.

\bibitem{Gent96B}
I.~P. Gent, P.~Prosser, and T.~Walsh, ``The constrainedness of search,'' in
  \emph{Proceedings of AAAI'96}, Portland, USA, Aug. 1999, pp. 246--252.

\bibitem{Bittle09}
S.~A. Bittle and M.~S. Fox, ``Learning and using hyper-heuristics for variable
  and value ordering in constraint satisfaction problems,'' in \emph{GECCO '09
  Proceedings of the 11th Annual Conference Companion on Genetic and
  Evolutionary Computation Conference: Late Breaking Papers}.\hskip 1em plus
  0.5em minus 0.4em\relax Quebec, Canada: ACM, Jul. 2009, pp. 2209--2212.

\bibitem{Boussemart04}
F.~Boussemart, F.~Hemery, C.~Lecoutre, and L.~Sais, ``Boosting systematic
  search by weighting constraints,'' in \emph{European Conference on Artificial
  Intelligence (ECAI'04)}, Valencia, Spain, Aug. 2004, pp. 146--150.

\bibitem{Bagnall201}
A.~J. Bagnall and G.~C. Cawley, ``On the use of default parameter settings in
  the empirical evaluation of classification algorithms,'' \emph{CoRR}, vol.
  abs/1703.06777, Mar. 2017.

\bibitem{bezdek2002vat}
J.~C. Bezdek and R.~J. Hathaway, ``{VAT: A tool for visual assessment of
  (cluster) tendency},'' in \emph{Proceedings of the 2002 International Joint
  Conference on Neural Networks}, vol.~3.\hskip 1em plus 0.5em minus
  0.4em\relax Honolulu, USA: IEEE, May 2002, pp. 2225--2230.

\bibitem{Pisinger2005}
D.~Pisinger, ``Where are the hard knapsack problems?'' \emph{Computers \&
  Operations Research}, vol.~32, no.~9, pp. 2271--2284, Sep. 2005.

\end{thebibliography}
\end{document}